\documentclass[10pt,twocolumn,letterpaper]{article}

\usepackage{cvpr}              
\usepackage[accsupp]{axessibility}
\definecolor{cvprblue}{rgb}{0.21,0.49,0.74}
\usepackage[breaklinks,colorlinks,allcolors=cvprblue]{hyperref}
\usepackage{multirow}
\usepackage[table]{xcolor}

\def\paperID{12210} 
\def\confName{CVPR}
\def\confYear{2026}

\title{Prompt-Anchored Vision–Text Distillation for Lifelong Person Re-identification}

\author{
Wen Wen$^{1,2,3}$ \quad Hao Chen$^{2}$\thanks{Corresponding author} \quad 
Shiliang Zhang$^{3}$\\
$^{1}$University of Electronic Science and Technology of China, China \\
$^{2}$Harbin Institute of Technology, Shenzhen, China \\
$^{3}$Peking University, China 
}

\begin{document}
\maketitle
\begin{abstract}
Lifelong person re-identification (LReID) aims to train a generalizable model with sequentially collected data. However, such models often suffer from semantic drift, limited adaptability, and catastrophic forgetting as new domains emerge. Existing exemplar-free approaches largely rely on visual-only distillation or parameter regularization, while overlooking the potential of auxiliary modalities, such as text, to preserve semantic stability and enable incremental plasticity. We observe that the frozen text encoder in pretrained vision–language models can serve as a stable semantic anchor across domains. To decouple the roles of vision and text, we propose Prompt-Anchored vision–text Distillation (PAD), an asymmetric vision–text framework for semantic alignment and cross-domain generalization. On the textual side, we distill prompts to preserve vision–text alignment under a fixed semantic space, acting as a global semantic reference rather than a dominant learning signal. On the visual side, an EMA-based teacher with an adaptive prompt pool enables domain-wise adaptation by allocating new slots while freezing past ones. Extensive experiments show that PAD substantially outperforms state-of-the-art methods across seen and unseen domains, achieving a strong balance between stability and plasticity. Project page: \url{https://github.com/zu-zi/PAD}.
\end{abstract}    
\section{Introduction}
\label{sec:intro}
Person re-identification (ReID) aims to retrieve the same individual across non-overlapping camera views, and has been extensively studied under static closed-world settings.
However, real-world surveillance systems continuously encounter data from evolving environments, new cameras, or changing time periods. In such scenarios, retraining models from scratch is impractical.
This dynamic scenario has motivated the study of lifelong person re-identification (LReID)~\cite{pu_cvpr2021,Wu2021GeneralisingWF}, where a model incrementally learns new identities while retaining previously acquired knowledge.

A key challenge of LReID lies in the catastrophic forgetting phenomenon: newly learned information often overwrites previously learned representations.
This problem becomes even more critical in person ReID, where large intra-person divergences (e.g., lighting, viewpoint, occlusion) coexist with small inter-person variations.
Such fine-grained differences make it difficult to preserve stable identity semantics over time.
Consequently, the model suffers from semantic drift, the learned embedding space gradually shifts, causing previously separable identities to become confused or misaligned.

\begin{figure}[t]
\centering
\includegraphics[width=1\linewidth]{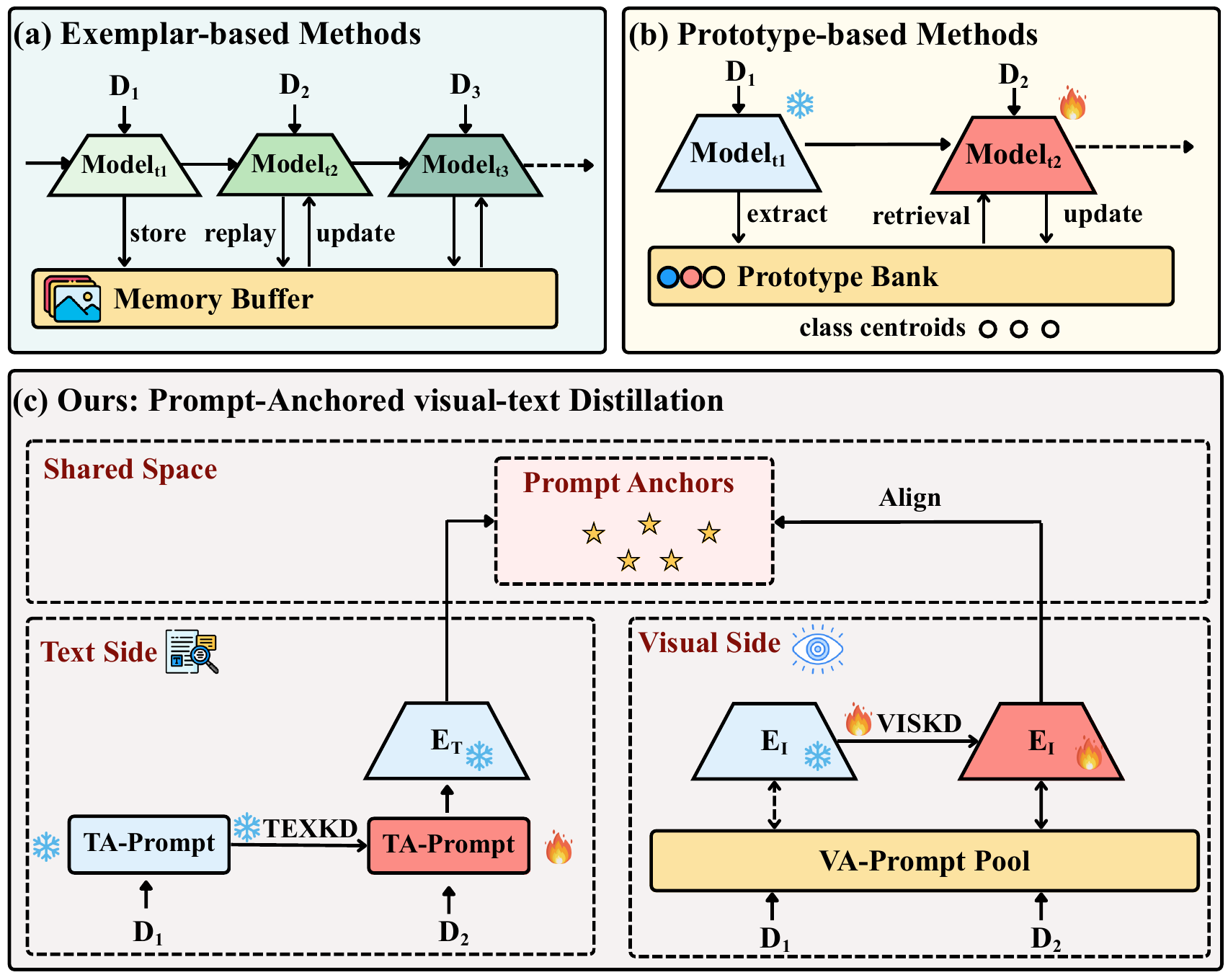}
\caption{
Comparison of lifelong ReID paradigms.
(a) Exemplar-based methods store raw samples for rehearsal. (b) Exemplar-free prototype-based methods store class centroids or distribution for visual knowledge distillation. 
(c) Our proposed PAD extends exemplar-free methods to both visual and text modalities:
a frozen text encoder provides stable semantic anchors,
while text distillation (TEXKD) and visual distillation (VISKD) respectively distill the TA-Prompt and visual branch,
achieving effective alignment in a shared vision–text space.
}
\label{fig:lreid_paradigms}
\end{figure}

To mitigate catastrophic forgetting, existing lifelong person ReID approaches can be roughly categorized into exemplar-based methods~\cite{yu2023lifelong,chen2025tpami} and exemplar-free prototype-based methods~\cite{xu2024lstkc,xu2024distribution}, as shown in Fig.~\ref{fig:lreid_paradigms}.
Exemplar-based methods store samples from previous domains to preserve past knowledge, which has proven to be effective but raises data storage and privacy concerns in real-world deployments.
To reduce the risk of privacy leakage, recent methods adopt visual knowledge distillation based on sample feature affinity~\cite{xu2024lstkc} or distribution statistics~\cite{xu2024distribution}, transferring knowledge without retaining old images.
However, these techniques operate solely within the visual modality and thus struggle to maintain the semantic consistency of person identities when domain distributions shift.
As a result, the learned feature space may remain locally discriminative but gradually drift away from the underlying identity semantics over time.

Meanwhile, the emergence of vision–language models such as CLIP~\cite{Radford2021LearningTV} and its variant CLIP-ReID~\cite{li2023clip} provides new opportunities for semantically grounded representation learning.
Textual information naturally offers domain-invariant semantic anchors—descriptions such as “a person wearing a red jacket” remain stable even when visual appearances vary due to pose or illumination changes.
This motivates us to exploit \emph{textual semantics as persistent anchors for guiding the evolution of visual representations} during exemplar-free lifelong learning.

In this paper, we propose \emph{Prompt-Anchored vision–text Distillation (PAD)}, a novel exemplar-free framework with an asymmetric vision–text design that introduces textual semantic guidance for visual feature learning in LReID.
Unlike prior works that rely on visual-only distillation or prompt-based adaptation in the visual space, PAD introduces a frozen textual semantic space as a global anchor while enabling adaptive visual learning.
We maintain a set of Text-Anchor Prompts (TA-Prompts) that align visual and textual representations, and a set of Visual-Adaptation Prompts (VA-Prompts) that facilitate incremental adaptation and memory across domains.
During each incremental stage, these prompts constrain the visual encoder to remain aligned with historical semantic knowledge.
This design effectively suppresses semantic drift while allowing flexible adaptation to new identities and domains.

Specifically, on the textual side, we freeze the pretrained text encoder to establish stable anchors as implicit semantic constraints, and train only the TA-Prompts to align visual and textual representations. In subsequent domains, a fixed teacher further performs distillation on TA-Prompts to prevent textual semantics from drifting away from previously learned domains.
On the visual side, we update the last few layers of the backbone, the classification head, and a visual prompt pool to balance plasticity and memory. An EMA-based teacher provides visual information distillation on the visual encoder, effectively mitigating fine-grained drift and enhancing cross-domain consistency.

In summary, our main contributions are threefold:
\begin{itemize}
    \item We introduce an exemplar-free LReID framework that uses a frozen textual semantic space as a global anchor, instead of relying solely on visual distillation.
    \item We design two prompts with asymmetric roles: TA-Prompt for semantic anchoring in the text space, and VA-Prompt for domain-wise adaptation in the visual space.
    \item Extensive experiments demonstrate that PAD achieves consistent improvements over prior work on both seen and unseen domains.
\end{itemize}

\section{Related Work}
\label{sec:related}

\noindent\textbf{CLIP-based person ReID.}
With the rapid progress of vision–language models, researchers have increasingly explored adapting CLIP for person ReID.
Li~\etal~\cite{li2023clip} develop CLIP-ReID, which integrates ambiguous textual descriptions with learnable tokens in a two-stage training paradigm, and has become a widely adopted baseline for person ReID.
Yu~\etal~\cite{yu2024tf} further present TF-CLIP, a one-stage framework that learns sequential prompts to capture temporal cues for video-based ReID.
MP-ReID~\cite{zhai2024multi} leverages both learnable prompts and large language model–generated prompts to enhance multi-modal representation learning.
Built upon textual inversion, PromptSG~\cite{yang2024pedestrian} introduces a lightweight inversion network to learn a pseudo-token representation for each individual, achieving efficient personalized modeling.

\noindent\textbf{Prompt-based lifelong learning.}
Prompt-based lifelong learning has gained increasing attention with the surge of pretrained models~\cite{dosovitskiy2020image,Radford2021LearningTV}, leveraging prompts to adapt frozen backbones to sequentially arriving data and novel classes.
L2P~\cite{wang2022learning} introduces a prompt pool to adapt a pretrained ViT for continual learning without revisiting previous data.
DualPrompt~\cite{wang2022dualprompt} extends L2P by separating prompts into G(eneral)-Prompt and E(xpert)-Prompt to balance generalization and specialization.
CODA-Prompt~\cite{smith2023coda} further develops a learnable prompt generator that dynamically constructs prompts from image features or patches.
Instead of maintaining an explicit pool or generator, CoSTEP~\cite{Zou_2025_CVPR} models the conditional distribution of prompts given input features, achieving adaptive prompt selection across tasks.
RainbowPrompt~\cite{hong2025rainbowprompt} introduces a diversity-aware prompt evolution mechanism that adaptively fuses prompts from multiple domains to mitigate forgetting.
Most of the above approaches focus on visual prompts, while recent studies have begun to explore multi-modal prompts for CLIP-based lifelong learning.
For instance, S-Prompt~\cite{wang2022s} and AttriCLIP~\cite{wang2023attriclip} enhance cross-modal alignment by jointly optimizing visual and textual prompts.
More recently, ENGINE~\cite{Zhou_2025_ICCV} integrates external knowledge from large language models to further improve the semantic consistency of CLIP-based lifelong learning.

\noindent\textbf{Lifelong person ReID.}
Lifelong person re-identification (LReID) aims to learn incrementally added person data while preserving previously acquired knowledge. Previous LReID methods can be roughly categorized into exemplar rehearsal methods and exemplar-free methods. Exemplar rehearsal methods~\cite{pu_cvpr2021,Wu2021GeneralisingWF,huang2022lifelong,chen2025tpami} store image samples in a memory buffer and conduct knowledge distillation or parameter regularization with stored samples. Despite the effectiveness of exemplar rehearsal, such methods may raise data storage and privacy concerns. Instead of raw samples, exemplar-free methods~\cite{xu2024lstkc,xu2024distribution,zhou2025distribution,xu2025self} store feature prototypes or distributions to regularize model parameter update in lifelong learning, which have less privacy concerns in real-world deployments. 

\noindent\textbf{Difference with related work.}
PAEMA~\cite{li2024exemplar} studies exemplar-free LReID with prompt-guided adaptation and EMA-based knowledge preservation, while DAFC~\cite{liu2025distribution} emphasizes distribution-aware forgetting compensation for exemplar-free LReID. DualPrompt~\cite{wang2022dualprompt} formulates continual learning with complementary prompts on a pretrained visual backbone, and ENGINE~\cite{Zhou_2025_ICCV} improves CLIP-based class-incremental learning by injecting external knowledge from both visual and textual modalities. CLIP-ReID~\cite{li2023clip} leverages textual semantics for static ReID, but does not consider continual adaptation.
Different from these works, PAD is built around a frozen textual semantic space that serves as a persistent anchor throughout lifelong learning. Our design is asymmetric: the textual branch is responsible for semantic preservation, while the visual branch remains adaptive through prompt expansion and distillation. Thus PAD is not a simple combination of prompting and distillation modules, but a vision--text formulation tailored to the stability--plasticity trade-off in LReID.
\section{Proposed Method}
\label{sec:method}
\begin{figure*}[t]
  \centering
  \includegraphics[width=1\linewidth]{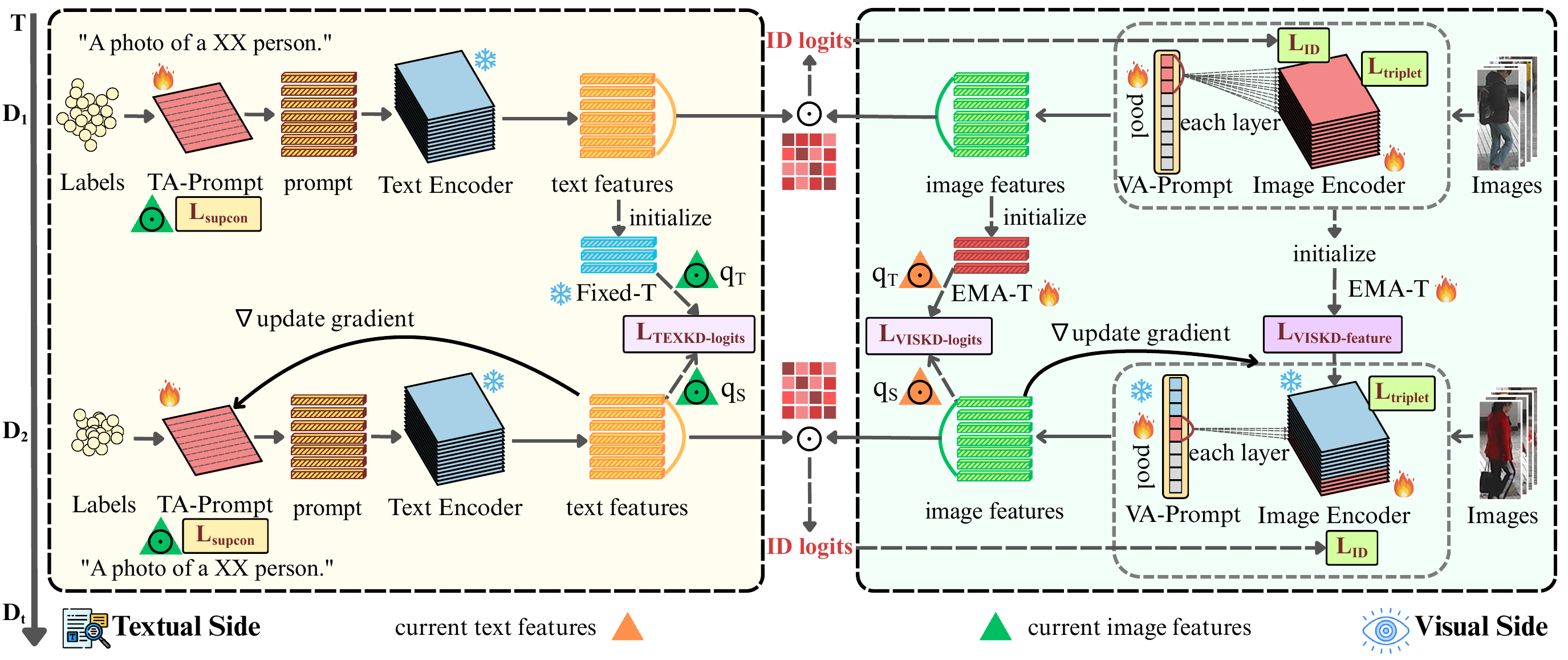}
  \caption{\textbf{Overview of the proposed PAD framework.}
  The framework consists of a textual branch (left) and a visual branch (right) that evolve across domains. On the textual side, we use a frozen text encoder and distill the learnable textual prompts (TA-Prompt). On the visual side, we construct a visual prompt (VA-Prompt) pool and train last layers of the image encoder with a two-term visual distillation loss. The textual branch provides semantic guidance during training, while only the image encoder is kept for inference.
  }
  \label{fig:pad-architecture}
\end{figure*}

\subsection{Formulation}
In the task of exemplar-free lifelong person re-identification (LReID), a stream of $T$ datasets denoted as $\mathcal{D}=\{D_t\}_{t=1}^{T}$ arrives sequentially for incremental training. Each dataset $D_t$ consists of a training set $D_t^{\text{train}}$ and a test set $D_t^{\text{test}}$, with no overlapping identities between them. During training, at step $t$, the model can only access the current training set $D_t^{\text{train}}$ for learning, while data from previous domains $\{D_i^{\text{train}}\}_{i < t}$ are inaccessible to avoid storage or privacy issues.

For evaluation, all seen-domain test sets $\{D_i^{\text{test}}\}_{i=1}^{T}$ are used to assess the model’s anti-forgetting and knowledge acquisition abilities. Additionally, a set of unseen datasets $\mathcal{D}^U=\{D_i\}_{i=1}^{U}$ is utilized to measure the generalization performance of the LReID model on unseen domains.

\subsection{Overview of PAD}
Conceptually inspired by CLIP-ReID~\cite{li2023clip}, the proposed Prompt-Anchored vision–text Distillation (PAD) framework extends cross-modal alignment into a lifelong setting.

As illustrated in Fig.~\ref{fig:pad-architecture}, PAD consists of a frozen text encoder, a partially trainable image encoder with only later layers unfrozen, and two prompt mechanisms that serve the textual and visual pathways, respectively. The two pathways play asymmetric roles: the textual branch preserves cross-domain semantics, while the visual branch remains adaptive to new domains.

In the textual pathway, the frozen text encoder provides a stable semantic coordinate system. The TA-Prompt is optimized with a weak textual distillation signal from a fixed text teacher. This weak alignment preserves long-term semantic consistency without over-constraining domain adaptation.

In the visual pathway, PAD employs a complementary form of strong distillation. A momentum-based EMA teacher regularizes high-level visual representations, preventing semantic drift as new domains accumulate. Meanwhile, the VA-Prompt operates inside the transformer backbone, allocating new prompt slots for each incoming domain while freezing previous ones, enabling continual visual adaptation while reducing forgetting of previously learned domains.

The overall training objective integrates contrastive alignment, identity classification, metric learning, and vision–text knowledge distillation (KD):
\begin{equation}
\mathcal{L}_{\mathrm{overall}}
= \mathcal{L}_{\mathrm{supcon}}
+ \mathcal{L}_{\mathrm{ID}}
+ \mathcal{L}_{\mathrm{triplet}}
+ \mathcal{L}_{\mathrm{KD}},
\label{eq:pad_main}
\end{equation}
where $\mathcal{L}_{\mathrm{supcon}}$ is a supervised contrastive loss~\cite{khosla2020supervised} that aligns TA-Prompt with the visual feature bank, while $\mathcal{L}_{\mathrm{ID}}$ and $\mathcal{L}_{\mathrm{triplet}}$ are logit-level cross-entropy loss and feature-level triplet loss~\cite{hermans2017defense}, respectively. The distillation term collects all textual and visual KD losses:
\begin{equation}
\mathcal{L}_{\mathrm{KD}}
= \lambda_{\text{text}} \mathcal{L}_{\text{TEXKD}}
+ \lambda_{\text{feat}} \mathcal{L}_{\text{featKD}}
+ \lambda_{\text{logit}} \mathcal{L}_{\text{logitKD}},
\label{eq:pad_kd}
\end{equation}
where $\mathcal{L}_{\text{TEXKD}}$ is the text-side logit KD that keeps TA-Prompt close to the frozen text anchor, while $\mathcal{L}_{\text{featKD}}$ and $\mathcal{L}_{\text{logitKD}}$ are the visual-side feature KD and logit KD that transfer knowledge from the EMA teacher to the partially trainable backbone. The coefficients $\lambda_{\text{text}}$, $\lambda_{\text{feat}}$, and $\lambda_{\text{logit}}$ control the contributions of each distillation pathway.

The text encoder supplies stable cross-domain semantic anchors during training, while only the image encoder is retained at inference, ensuring high computational efficiency.

\subsection{Textual Side}
PAD establishes a dual-level semantic alignment on the textual side, combining implicit anchoring from the frozen text encoder with explicit guidance through distillation of the learnable prompts. The implicit alignment provides the primary semantic anchoring, while explicit TEXKD only offers lightweight regularization for long-term consistency.

\subsubsection{Implicit alignment}
The frozen text encoder inherited from the pretrained vision–language model defines a stable linguistic coordinate system that remains invariant across domains. Within this space, the TA-Prompt learns a set of trainable prompt tokens~\cite{li2023clip} that generate class-level textual embeddings, effectively bridging the mapping between textual category descriptions and visual representations.

Concretely, for each identity $y_i$ we extract its visual feature 
$\mathbf{v}_i \in \mathbb{R}^d$ from the current image encoder, and obtain the corresponding textual feature 
$\mathbf{t}_i \in \mathbb{R}^d$ by inserting the class-specific prompt tokens generated by TA-Prompt into the fixed text template and feeding the resulting sentence into the frozen text encoder. 
Both features are $\ell_2$-normalized, and TA-Prompt is optimized with a symmetric image--text SupCon loss~\cite{khosla2020supervised}:
\begin{equation}
\mathcal{L}_{\mathrm{supcon}}
= \mathrm{SupCon}(\mathbf{v}\!\to\!\mathbf{t})
+ \mathrm{SupCon}(\mathbf{t}\!\to\!\mathbf{v}),
\label{eq:supcon}
\end{equation}
where $\mathrm{SupCon}(\mathbf{v}\!\to\!\mathbf{t})$ and 
$\mathrm{SupCon}(\mathbf{t}\!\to\!\mathbf{v})$ are supervised contrastive losses computed over 
$\{\mathbf{v}_i\}$ and $\{\mathbf{t}_i\}$ with identity labels $\{y_i\}$. Positive pairs share the same identity label, and negative pairs come from different identities.

Because the text encoder is frozen, optimizing $\mathcal{L}_{\mathrm{supcon}}$ only updates the prompt parameters while keeping them aligned with the stable semantic anchors. In other words, each domain adapts its TA-Prompt to better match the current visual distribution, yet all domains remain tied to a unified cross-modal semantic space dictated by the frozen text backbone. This produces an implicit alignment effect: visual clusters are consistently pulled toward a set of class-level textual anchors shared across domains, even without any explicit text-side distillation. This implicit SupCon-driven alignment already acts as a strong, domain-agnostic semantic regularizer, explaining the robustness of our baseline when explicit TEXKD is disabled.

\subsubsection{Explicit alignment}
To obtain finer semantic consistency without sacrificing plasticity, PAD adds a deliberately weak distillation term on top of the implicit image--text contrastive loss.

For each domain, we first construct a frozen text teacher by loading the previous checkpoint and keeping only its textual branch fixed. Passing all identity labels through this teacher yields a global text bank $t^{\mathrm{tea}} \in \mathbb{R}^{C \times d}$, which is cached once and reused throughout training. During optimization, we sample a batch of cached visual features $v \in \mathbb{R}^{B \times d}$ from the current domain, and select a label subset consisting of the batch identities and a set of randomly sampled negative identities. For this subset, the student TA-Prompt produces text embeddings $t^{\mathrm{stu}}$, while the teacher provides the corresponding anchors from $t^{\mathrm{tea}}$. We define a temperature-scaled softmax cosine similarity function:
\begin{equation}
q(v,t,\tau,\gamma) = 
\frac{\exp{\left(\gamma\, (v \cdot t_{+}) / \tau\right)}}
     {\sum\nolimits_{i=1}^{K}\exp{\left(\gamma\, (v \cdot t_{i}) / \tau\right)}},
\label{eq:textkd_logits}
\end{equation}
where $K$ denotes the size of the sampled identity subset, including the batch identities and the randomly sampled negative identities. $v \cdot t$ is the cosine similarity between visual and text embeddings. $\tau$ is a temperature parameter that smooths the similarity distribution, and $\gamma$ is a learnable logit-scaling factor that adjusts the overall sharpness of similarity scores.

To softly align these two distributions, PAD applies a temperature-scaled textual distillation loss,
denoted $\mathcal{L}_{\mathrm{TEXKD}}$:
\begin{equation}
\mathcal{L}_{\mathrm{TEXKD}}
= \tau^2\,
D_{\mathrm{KL}}\!\left( q(v,t^{\mathrm{tea}},\tau,\gamma) \,\middle\|\, q(v,t^{\mathrm{stu}},\tau,\gamma) \right),
\label{eq:explicit_texkd}
\end{equation}
where $D_{\mathrm{KL}}$ denotes the Kullback--Leibler divergence between the teacher and student similarity distributions.

In practice, we keep the teacher fixed within a domain and do \emph{not} apply EMA on the textual branch: because the text encoder is fixed and only the prompts are updated, an EMA teacher would quickly collapse to the student, causing the KL term to vanish after only a few iterations. Using a frozen teacher avoids this collapse and provides a stable reference drawn from previous domains.

This explicit alignment therefore complements the symmetric image--text contrastive loss: while contrastive learning adapts TA-Prompt to the current visual distribution, $\mathcal{L}_{\mathrm{TEXKD}}$ gently keeps its image-conditioned class scores compatible with the historical text bank. With a small distillation weight, PAD preserves the plasticity of the textual side across domains while still leveraging the teacher as a semantic anchor accumulated along the lifelong sequence. We intentionally keep $\mathcal{L}_{\mathrm{TEXKD}}$ weak, since overly strong text-side regularization would over-constrain prompt adaptation on new domains.

\subsection{Visual Side}
\subsubsection{Visual distillation}
To regulate the adaptation of the partially unfrozen visual backbone and prevent semantic drift across domains, PAD applies an EMA-based visual distillation strategy. A momentum teacher is maintained by updating its parameters after every iteration:
\begin{equation}
\theta_{tea} \leftarrow \alpha \theta_{tea} + (1 - \alpha)\theta_{stu},
\end{equation}
where $\theta_{stu}$ and $\theta_{tea}$ denote the student and teacher parameters of the visual branch, and $\alpha \in (0, 1)$ is the momentum coefficient. This EMA teacher provides a temporally smoothed target that reflects the recent training trajectory while filtering out noisy fluctuations, stabilizing continual updates under domain shift.

Given an input batch of images, we compute a triplet of visual representations from both the student and the EMA teacher:
\[
v^{stu}_i = \{ v^{stu}_{11}, v^{stu}_{12}, v^{stu}_{proj} \}, \qquad
v^{tea}_i = \{ v^{tea}_{11}, v^{tea}_{12}, v^{tea}_{proj} \},
\]
corresponding to the penultimate-layer token output, the final-layer token output, and the final projected embedding of the visual transformer. These multi-level features are distilled via a mean-squared regression:
\begin{equation}
\mathcal{L}_{\mathrm{featKD}}
= \frac{1}{3} \sum_{i=1}^{3} \left\| v^{stu}_i - v^{tea}_i \right\|_2^2.
\end{equation}

In parallel, we apply a logit-level distillation that mirrors the textual branch but is anchored on the domain-specific text feature bank. Concretely, both the student image features and its EMA teacher counterparts are projected onto the same fixed textual embedding set, producing two vision–text similarity distributions through plain inner-product scoring without learnable logit-scaling factor. A temperature-smoothed KL divergence is then used to match these two distributions:
\begin{equation}
\mathcal{L}_{\mathrm{logitKD}}
= \tau^2\,
D_{\mathrm{KL}}\!\left( q(v^{\mathrm{tea}},t,\tau) \,\middle\|\, q(v^{\mathrm{stu}},t,\tau) \right).
\end{equation}

The overall visual distillation term is therefore
\begin{equation}
\mathcal{L}_{\mathrm{VISKD}}
= \lambda_{\text{feat}} \mathcal{L}_{\mathrm{featKD}}
+ \lambda_{\text{logit}} \mathcal{L}_{\mathrm{logitKD}},
\end{equation}
with $\lambda_{\text{feat}}$ and $\lambda_{\text{logit}}$ controlling the respective strengths. Because visual representations sit at a finer granularity and are highly sensitive to domain distribution shifts, PAD applies visual distillation with a relatively strong weight. Combined with selectively unfrozen late transformer blocks, this provides sufficient plasticity for learning new domains, while using the EMA teacher as an anchor to prevent the learned feature space from collapsing or drifting away from previous knowledge.

\subsubsection{VA-Prompt design}
The VA-Prompt is inspired by the design philosophy of DualPrompt~\cite{wang2022dualprompt} and consists of two complementary components: a General Prompt (G-Prompt) and an Expert Prompt (E-Prompt).
For each transformer layer in the image encoder, the G-Prompt provides a set of globally shared tokens, while the E-Prompt maintains a pool of expert-specific prompts. 
During forward propagation, the current input representation serves as a query to select the top-$k$ most relevant experts based on cosine similarity. 
The selected prompts are concatenated after the [CLS] token and before the patch tokens, forming an extended token sequence $[\text{CLS}, G, E, \text{patches}]$.
After the transformer block processes this sequence, the inserted prompts are stripped off, ensuring that no residual tokens overwrite the original structure of the visual representation.

Across domains, each new domain activates a fresh set of prompt slots, while freezing those from previous domains. This yields a dynamic prompt pool that grows without interfering with previously acquired knowledge.

\subsubsection{Selective layer unfreezing}
Before entering the lifelong sequence, the model first adapts the pretrained CLIP backbone to the ReID task on the initial domain, without distillation or layer freezing.
This step shifts the pretrained model from generic feature learning to ReID-specific representation learning.
In subsequent domains, prompt-based adaptation cannot accommodate large-scale unfreezing, while conventional distillation methods usually depend on full or deep-layer fine-tuning.

PAD adopts a balanced strategy: it only unfreezes the last few transformer blocks and the classification heads—where forgetting and transfer are most pronounced, while keeping lower layers frozen to preserve domain-invariant representations.
This selective unfreezing strategy mitigates computational cost and over-constraint, yet still enables effective adaptation.
\section{Experimental Results}
\label{sec:experiment}

\subsection{Datasets and Evaluation Metrics}
We conduct experiments across a total of twelve domains to comprehensively assess the lifelong adaptation and generalization ability of PAD.
The seen domains include Market1501~\cite{Zheng2015ScalablePR}, CUHK-SYSU~\cite{Xiao2017JointDA}, DukeMTMC-reID~\cite{ristani2016MTMC}, MSMT17~\cite{wei2018person}, and CUHK03~\cite{Li2014DeepReIDDF}, while the unseen domains consist of CUHK01~\cite{Li2012HumanRW}, CUHK02~\cite{Li2013LocallyAF}, VIPeR~\cite{Gray2008ViewpointIP}, PRID2011~\cite{hirzer11}, i-LIDS~\cite{Zheng2009AssociatingGO}, GRID~\cite{Loy2009MulticameraAC}, and SenseReID~\cite{Zhao2017SpindleNP}. Following the community's mainstream lifelong ReID protocol, we adopt the All-Known-All (AKA)~\cite{pu_cvpr2021} benchmarks, \ie
AKA-order1 and AKA-order2, which define the sequential training and evaluation orders.

Each domain is introduced incrementally without any replay of past data, and the model is evaluated on both the current and previously seen domains after each training stage, as well as on all unseen domains for cross-domain generalization. All datasets follow their standard splits, and performance is measured by mean Average Precision (mAP) and Cumulative Matching Characteristic at Rank-1 (R1). Our evaluation follows established lifelong ReID benchmarks and enables direct comparison with prior work. Additional evaluation under the LPW-s2~\cite{xu2025self} substitution protocol is provided in the supplementary material.

\subsection{Implementation Details}
We adopt CLIP ViT-B/16 as the visual backbone and follow the standard lifelong protocol over five domains. Images are resized to $256\times128$, augmented with random flipping and erasing, and sampled with an identity-balanced sampler. All models are trained using Adam with a batch size of 64 on a single NVIDIA RTX~3090 GPU.

On the textual side, we freeze the CLIP text encoder and only optimize TA-Prompt, implemented as class-specific contextual tokens. Text-side distillation applies only logit-level KL loss with weight $\lambda_{\text{text}}=0.5$, temperature $\tau=0.07$, and the learnable scaling factor $\gamma$ initialized to $7.0$.

VA-Prompt adopts 6 general + 6 expert tokens per layer (pool size 36, Top-K=4). New expert slots are activated per domain, while all previous slots are frozen. Only the last few transformer blocks and the classification head are unfrozen; the backbone uses a base LR of $5\times10^{-6}$. Visual distillation employs an EMA teacher with momentum $\alpha=0.997$, combined with feature-level MSE and logit-level KL losses weighted by $\lambda_{\text{feat}}=\lambda_{\text{logit}}=0.5$, $\tau=4.0$.

\subsection{Comparison with State-of-the-art Methods}
We compare PAD with representative state-of-the-art methods, including generic lifelong methods~\cite{li2017learning,wang2022dualprompt,li2024fcs} and ReID-specific lifelong methods~\cite{pu_cvpr2021,sun2022patch,xu2024lstkc,cui2024learning,xu2024distribution,li2024exemplar,zhou2025distribution,xu2025dask,liu2025distribution}. The results are summarized in Table~\ref{tab:aka_order1} and Table~\ref{tab:aka_order2}, corresponding to AKA-order1 and AKA-order2, respectively. Over 7 random seeds on AKA-order1, PAD achieves stable final-stage performance (seen-domain average: 70.30$\pm$0.49 mAP and 80.98$\pm$0.31 R1). PAD achieves the best overall averages on both seen and unseen domains under AKA-order1 and AKA-order2.

\begin{figure}[t]
    \centering
    \begin{subfigure}[t]{0.49\linewidth}
        \centering
        \includegraphics[width=\linewidth]{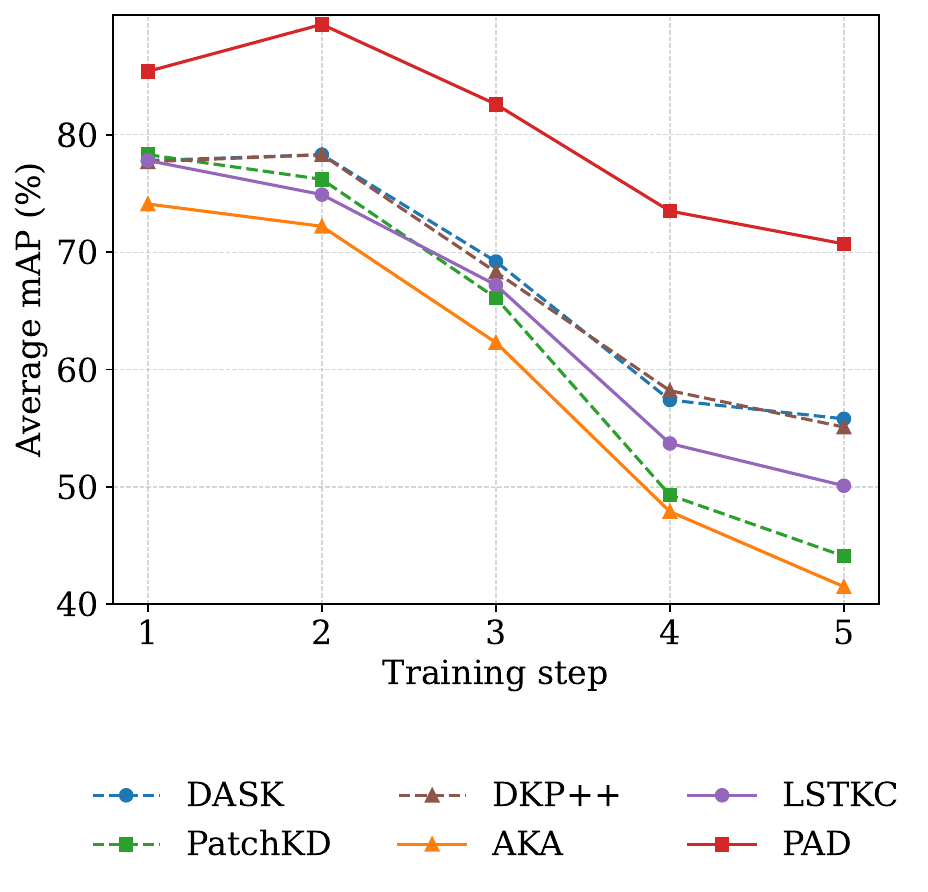}
    \end{subfigure}
    \hfill
    \begin{subfigure}[t]{0.49\linewidth}
        \centering
        \includegraphics[width=\linewidth]{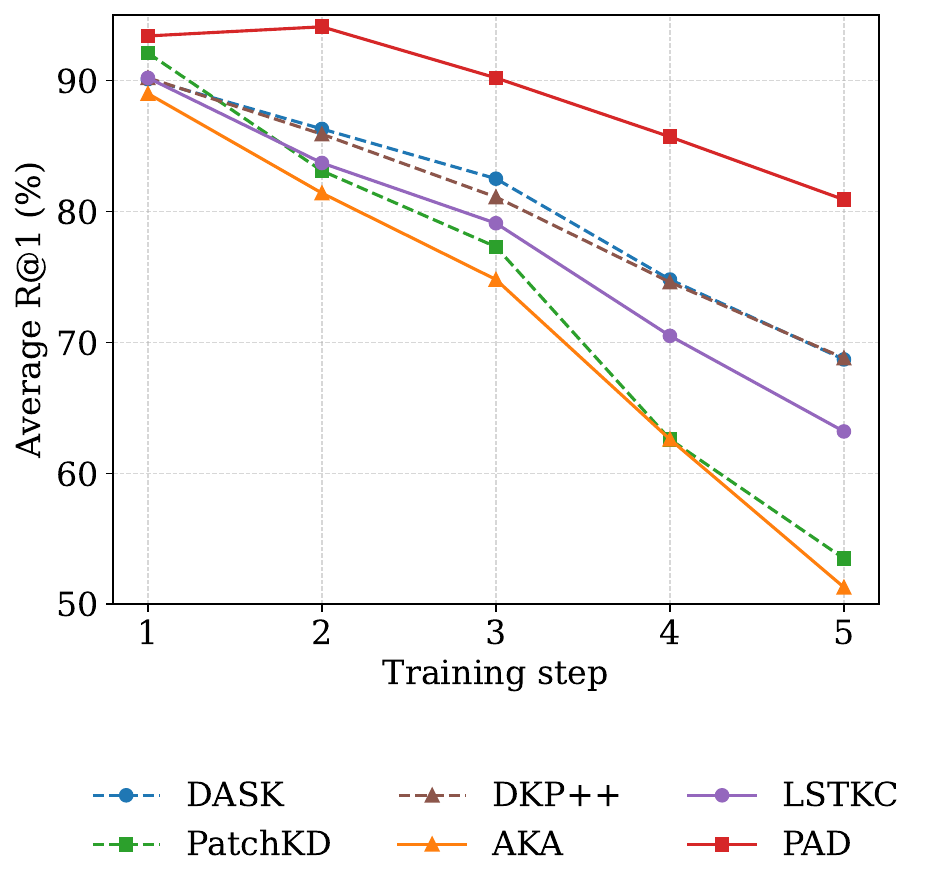}
    \end{subfigure}

    \vspace{1mm}
    \caption{
        \textbf{Performance tendency on seen domains (AKA-order1).}
        After each training step, the model is evaluated on the already-seen domains.
    }
    \label{fig:trend_seen}
\end{figure}

\begin{figure}[t]
    \centering
    \begin{subfigure}[t]{0.49\linewidth}
        \centering
        \includegraphics[width=\linewidth]{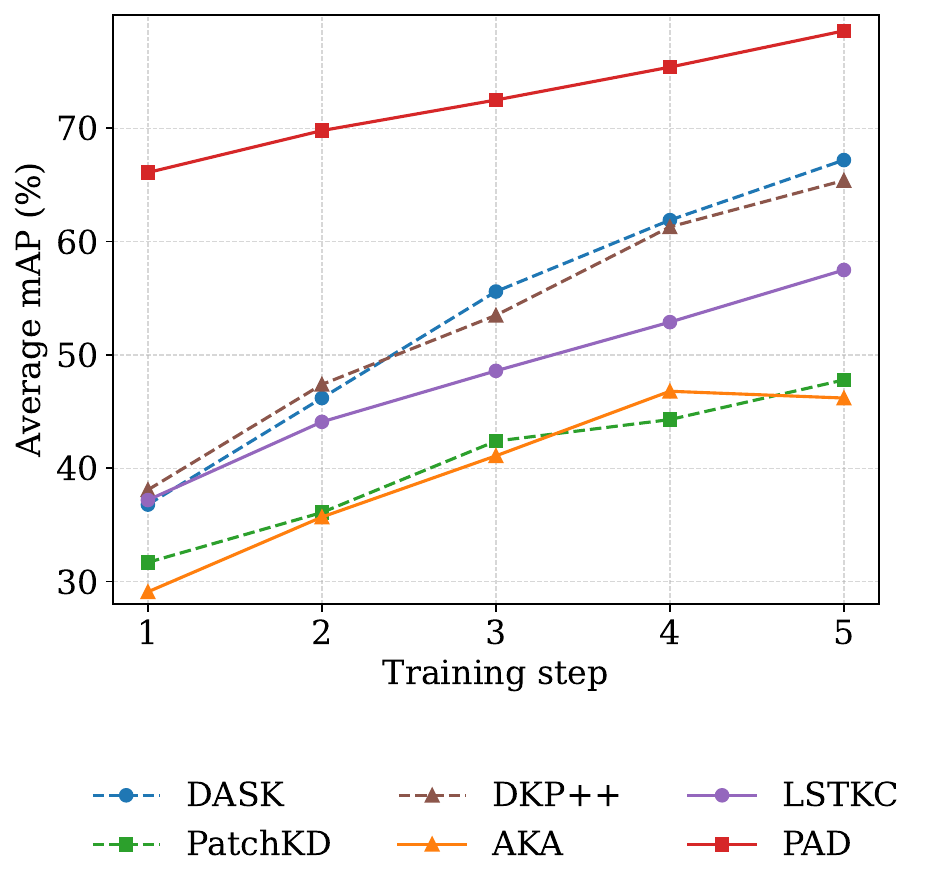}
    \end{subfigure}
    \hfill
    \begin{subfigure}[t]{0.49\linewidth}
        \centering
        \includegraphics[width=\linewidth]{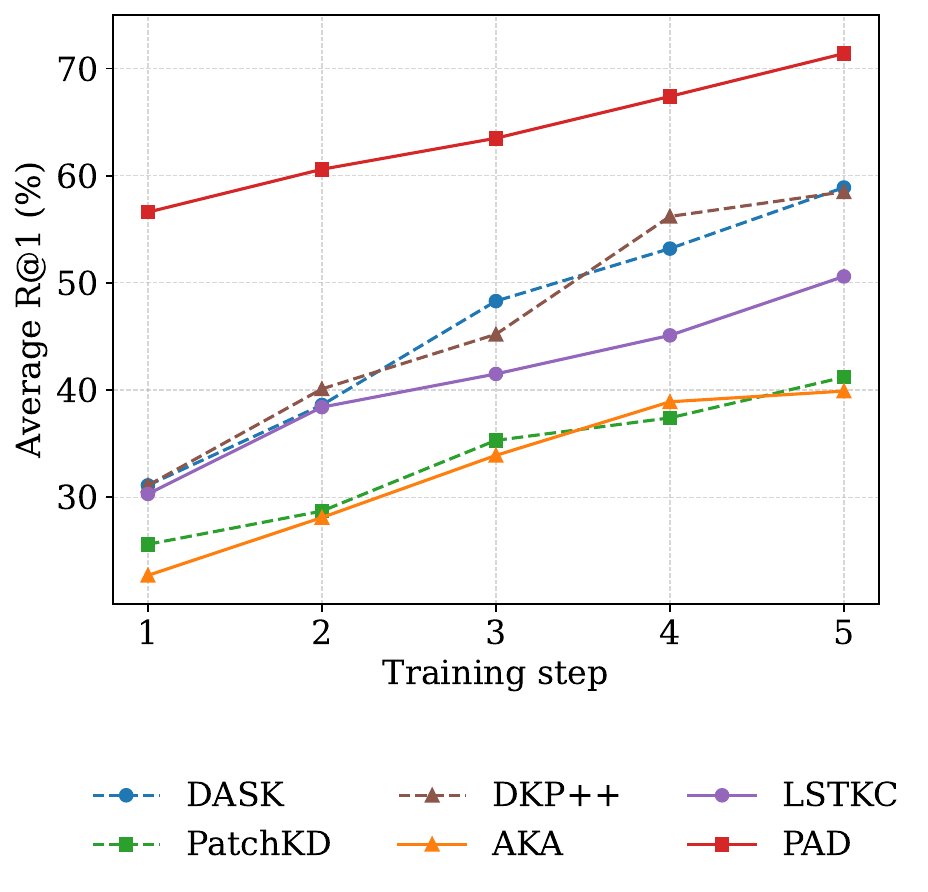}
    \end{subfigure}

    \vspace{1mm}
    \caption{
        \textbf{Performance tendency on unseen domains (AKA-order1).}
        After each training step, the performance of all unseen domains is evaluated.
    }
    \label{fig:trend_unseen}
\end{figure}

To provide additional intuition about PAD's sequential behavior, we further visualize the stage-wise performance tendency on AKA-order1. As shown in Fig.~\ref{fig:trend_seen}, PAD maintains stable performance on previously seen domains throughout the sequence. Fig.~\ref{fig:trend_unseen} further shows that unseen-domain performance improves as more tasks are introduced.

\begin{table*}[t]\small
	\centering
	\setlength{\tabcolsep}{4pt}	
	\caption{Performance comparison with state-of-the-art methods on AKA-order1. 
	The optimal and suboptimal values are highlighted in red and blue, respectively. 
	AKA-order1 is Market-1501$\to$CUHK-SYSU$\to$ DukeMTMC$\to$MSMT17$\to$CUHK03.}
	\label{tab:aka_order1}
	\begin{tabular}{l|c|cc|cc|cc|cc|cc|cc|cc}
		\hline
		\multirow{2}{*}{Method}&\multirow{2}{*}{Venue}&
		\multicolumn{2}{c|}{Market-1501}&\multicolumn{2}{c|}{CUHK-SYSU}&\multicolumn{2}{c|}{DukeMTMC}&\multicolumn{2}{c|}{MSMT17}&
		\multicolumn{2}{c|}{CUHK03}&\multicolumn{2}{c|}{\textbf{Seen-Avg}}&\multicolumn{2}{c}{\textbf{Unseen-Avg}}\\
		\cline{3-16}
		&&mAP &R1 &mAP &R1 &mAP &R1 &mAP &R1 &mAP &R1 &mAP &R1 &mAP &R1\\
		\hline
		LwF~\cite{li2017learning}&PAMI'17&56.3 &77.1 &72.9 &75.1 &29.6 &46.5 &6.0 &16.6 &36.1 &37.5 &40.2 &50.6 &47.2 &42.6\\
        DualPrompt~\cite{wang2022dualprompt}&ECCV'22&62.3&82.5&70.0&75.0&40.5&61.0&8.2&28.1&35.6&35.9&43.3&56.5&41.4&36.8\\
        FCS~\cite{li2024fcs}&CVPR'24&58.3&78.5&75.1&76.2&42.6&59.8&10.2&24.3&35.3&34.9&44.3&54.7&52.1&44.2\\
        \hline
		AKA~\cite{pu_cvpr2021}&CVPR'21 &58.1 &77.4 &72.5 &74.8 &28.7 &45.2 &6.1 &16.2 &38.7 &40.4 &40.8 &50.8 &47.6 &42.6 \\
		PatchKD~\cite{sun2022patch}& MM'22&68.5 &85.7 &75.6 &78.6 &33.8 &50.4 &6.5 &17.0 &34.1 &36.8 &43.7 &53.7 &49.1 &45.4\\
		LSTKC~\cite{xu2024lstkc}&AAAI'24 &54.7 &76.0 &81.1 &83.4 &49.4 &66.2 &20.0 &43.2 &44.7 &46.5 &50.0 &63.1 &57.0 &49.9\\
		C2R~\cite{cui2024learning}&CVPR'24 &69.0 &86.8 &76.7 &79.5 &33.2 &48.6 &6.6 &17.4 &35.6 &36.2 &44.2 &53.7 &48.7 &41.9\\
		DKP~\cite{xu2024distribution}&CVPR'24 &60.3 &80.6 &83.6 &85.4 &51.6 &68.4 &19.7 &41.8 &43.6 &44.2 &51.8 &64.1 &59.2 &51.6\\
        PAEMA~\cite{li2024exemplar}&IJCV'24&71.7&86.9&\textcolor{blue}{\textbf{90.0}}&\textcolor{blue}{\textbf{91.3}}&\textcolor{red}{\textbf{66.0}}&\textcolor{red}{\textbf{79.0}}&\textcolor{blue}{\textbf{31.4}}&55.3&50.0&50.8&61.8&72.7&\textcolor{blue}{\textbf{70.3}}&\textcolor{blue}{\textbf{63.2}}\\
        DKP++~\cite{zhou2025distribution}&PAMI'25 &63.9 &83.6 &83.7 &85.3 &56.7 &74.1 &26.5 &52.6 &46.5 &47.6 &55.4 &68.6 &65.7 &58.5\\
		DASK~\cite{xu2025dask}&AAAI'25 &61.2 &82.3 &81.9 &83.7 
		&58.5 &74.6 &29.1 &\textcolor{blue}{\textbf{57.6}} &46.2 &48.1 &55.4 &69.3 &65.3 &58.4\\
		DAFC~\cite{liu2025distribution}&Arxiv'25 &\textcolor{red}{\textbf{81.7}}&\textcolor{blue}{\textbf{90.6}}&89.7&90.9&64.5&\textcolor{blue}{\textbf{78.9}} &27.1 &52.9 &\textcolor{blue}{\textbf{64.0}} &\textcolor{blue}{\textbf{66.1}} &\textcolor{blue}{\textbf{65.6}} &\textcolor{blue}{\textbf{75.9}}&-&-\\	
        \hline
        \textbf{PAD (Ours)} &
        & \textcolor{blue}{\textbf{81.2}} & \textcolor{red}{\textbf{92.0}}   
        & \textcolor{red}{\textbf{92.6}} & \textcolor{red}{\textbf{93.5}}   
        & \textcolor{blue}{\textbf{64.9}} & 77.8   
        & \textcolor{red}{\textbf{46.9}} & \textcolor{red}{\textbf{72.0}}   
        & \textcolor{red}{\textbf{68.1}} & \textcolor{red}{\textbf{69.5}}   
        & \textcolor{red}{\textbf{70.7}} & \textcolor{red}{\textbf{81.0}}   
        & \textcolor{red}{\textbf{78.6}} & \textcolor{red}{\textbf{71.4}}   
        \\
		\hline
	\end{tabular}
\end{table*}

\begin{table*}[t]\small
	\centering
	\setlength{\tabcolsep}{4pt}
	\caption{Performance comparison with state-of-the-art methods on AKA-order2. The optimal and suboptimal values are highlighted in red and blue, respectively. AKA-order2 is DukeMTMC$\to$MSMT17$\to$Market-1501$\to$ CUHK-SYSU$\to$ CUHK03.}
	\label{tab:aka_order2}
	\begin{tabular}{l|c|cc|cc|cc|cc|cc|cc|cc}
		\hline
		\multirow{2}{*}{Method}&\multirow{2}{*}{Venue}&
		\multicolumn{2}{c|}{DukeMTMC}&\multicolumn{2}{c|}{MSMT17}&\multicolumn{2}{c|}{Market-1501}&\multicolumn{2}{c|}{CUHK-SYSU}& \multicolumn{2}{c|}{CUHK03}&\multicolumn{2}{c|}{\textbf{Seen-Avg}}&\multicolumn{2}{c}{\textbf{Unseen-Avg}}\\
		\cline{3-16}
		&&mAP &R1 &mAP &R1 &mAP &R1 &mAP &R1 &mAP &R1 &mAP &R1 &mAP &R1\cr 
		\hline
		LwF~\cite{li2017learning}&PAMI'17 &42.7 &61.7 &5.1 &14.3 &34.4 &58.6 &70.0 &73.0 &34.1 &34.1 &37.2 &48.4 &44.0 &40.1 \\
        FCS~\cite{li2024fcs}&CVPR'24 &53.6&70.0&9.5&23.5&48.7&69.9&76.2&78.2&37.1&38.4&45.0&56.0&52.7&45.1\\
        \hline
		AKA~\cite{pu_cvpr2021}&CVPR'21 &42.2 &60.1 &5.4 &15.1 &37.2 &59.8 &71.2 &73.9 &36.9 &37.9 &38.6 &49.4 &46.0 &41.7\\
		PatchKD~\cite{sun2022patch}&MM'22 &58.3 &74.1 &6.4 &17.4 &43.2 &67.4 &74.5 &76.9 &33.7 &34.8 &43.2 &54.1 &48.6 &44.1\\
		LSTKC~\cite{xu2024lstkc}&AAAI'24 &49.9 &67.6 &14.6 &34.0 &55.1 &76.7 &82.3 &83.8 &46.3 &48.1 &49.6 &62.1 &57.6 &49.6\\		
		C2R~\cite{cui2024learning}&CVPR'24 &59.7&75.0&7.3 &19.2 &42.4 &66.5 &76.0 &77.8 &37.8 &39.3 &44.7 &55.6 &48.5 &41.7\\
		DKP~\cite{xu2024distribution}&CVPR'24 &53.4 &70.5 &14.5 &33.3 &60.6 &81.0 &83.0 &84.9 &45.0 &46.1 &51.3 &63.2 &59.0 &51.6 \\
        PAEMA~\cite{li2024exemplar}&IJCV'24&67.2&79.8&26.0&49.4&69.8&85.8&\textcolor{blue}{\textbf{91.0}}&\textcolor{blue}{\textbf{89.9}}&49.3&49.7&60.4&71.1&\textcolor{blue}{\textbf{69.4}}&\textcolor{blue}{\textbf{62.5}}\\
        DKP++~\cite{zhou2025distribution}&PAMI'25&55.7&73.9&22.2&47.8&65.9&85.6&84.4&86.5&49.5&51.7&55.5&69.1&64.9&57.9\\
		DASK~\cite{xu2025dask}&AAAI'25 &55.7 &74.4 &25.2 &51.9 &71.6&\textcolor{blue}{\textbf{87.7}}&84.8&86.2&48.4 &49.8 &57.1&70.0&65.5 &57.9\\
		DAFC~\cite{liu2025distribution}&Arxiv'25 &\textcolor{red}{\textbf{71.1}} &\textcolor{red}{\textbf{83.2}}&\textcolor{blue}{\textbf{31.2}}&\textcolor{blue}{\textbf{58.0}}&\textcolor{blue}{\textbf{72.2}}&87.4&88.2 &89.3&\textcolor{blue}{\textbf{60.6}}&\textcolor{blue}{\textbf{62.9}}&\textcolor{blue}{\textbf{64.7}}&\textcolor{blue}{\textbf{76.2}}&-&-\\
		\hline
        \textbf{PAD (Ours)}&
        & \textcolor{red}{\textbf{71.1}} & \textcolor{blue}{\textbf{82.6}}   
        & \textcolor{red}{\textbf{37.7}} & \textcolor{red}{\textbf{64.0}}   
        & \textcolor{red}{\textbf{77.5}} & \textcolor{red}{\textbf{89.8}}   
        & \textcolor{red}{\textbf{93.9}} & \textcolor{red}{\textbf{94.9}}   
        & \textcolor{red}{\textbf{66.4}} & \textcolor{red}{\textbf{68.5}}   
        & \textcolor{red}{\textbf{69.3}} & \textcolor{red}{\textbf{80.0}}   
        & \textcolor{red}{\textbf{76.2}} & \textcolor{red}{\textbf{68.6}}   
        \\
        \hline
	\end{tabular}
\end{table*}

\subsection{Ablation Studies}
To further verify the effectiveness of each component in the proposed PAD framework, we conduct comprehensive ablation studies under the AKA-order1 benchmark. All experiments follow the same lifelong training protocol.

\subsubsection{Component Analysis}
We begin by incrementally assembling the proposed framework to assess the contribution of each module.
Starting from a fully fine-tuned CLIP-ReID~\cite{li2023clip} baseline without any prompt or distillation, we progressively introduce the freezing scheme, VA-Prompt, textual distillation, and visual distillation.

\begin{table*}[t]
\centering
\small
\caption{Ablation study on AKA-order1. Columns indicate modules: \textbf{Freeze}—PAD freezing scheme, \textbf{VA}—Visual Adaptive Prompt, \textbf{TEXKD}—textual fixed distillation, \textbf{VISKD}—visual EMA distillation.}
\label{tab:component}
\setlength{\tabcolsep}{4.2pt}
\renewcommand{\arraystretch}{1.05}
\begin{tabular}{lcccc|cc|cc|cc|cc|cc|cc}
\hline
\multirow{2}{*}{ID} &
\multirow{2}{*}{Freeze} &
\multirow{2}{*}{VA} &
\multirow{2}{*}{TEXKD} &
\multirow{2}{*}{VISKD} &
\multicolumn{2}{c|}{Market1501} &
\multicolumn{2}{c|}{CUHK-SYSU} &
\multicolumn{2}{c|}{DukeMTMC} &
\multicolumn{2}{c|}{MSMT17} &
\multicolumn{2}{c|}{CUHK03} &
\multicolumn{2}{c}{Seen Avg} \\
\cline{6-17}
 &  &  &  &  & mAP & R1 & mAP & R1 & mAP & R1 & mAP & R1 & mAP & R1 & mAP & R1 \\
\hline
S0 &  &  &  &  & 67.2 & 83.6 & 84.1 & 85.8 & 58.8 & 74.3 & 41.3 & 67.2 & \textbf{79.4} & \textbf{81.4} & 66.2 & 78.5 \\
S1 & \checkmark &  &  &  & \textbf{81.9} & \textbf{92.8} & 90.7 & 91.7 & 58.6 & 75.1 & 40.8 & 67.9 & 62.1 & 64.2 & 66.8 & 78.4 \\
S2 & \checkmark & \checkmark &  &  & 78.2 & 90.7 & 90.5 & 91.8 & 59.7 & 76.4 & 42.8 & 69.5 & 69.6 & 72.0 & 68.2 & 80.1 \\
S3 & \checkmark & \checkmark & \checkmark &  & 79.1 & 91.0 & 90.9 & 92.0 & 59.0 & 75.5 & 43.4 & 70.1 & 69.7 & 73.2 & 68.4 & 80.4 \\
S4 & \checkmark & \checkmark &  & \checkmark & 79.2 & 91.1 & 92.3 & 93.3 & 64.1 & 77.3 & 45.6 & 71.2 & 68.1 & 69.1 & 69.9 & 80.4 \\
S5 & \checkmark & \checkmark & \checkmark & \checkmark & 81.2 & 92.0 & \textbf{92.6} & \textbf{93.5} & \textbf{64.9} & \textbf{77.8} & \textbf{46.9} & \textbf{72.0} & 68.1 & 69.5 & \textbf{70.7} & \textbf{81.0} \\
\hline
\end{tabular}
\end{table*}

Table~\ref{tab:component} reflects the expected stability--plasticity trade-off. The fully fine-tuned baseline (S0) adapts well to later domains but suffers stronger forgetting on earlier ones, while the freezing-only variant (S1) favors early-domain retention at the cost of adaptation. Adding VA-Prompt (S2) restores plasticity, and TEXKD (S3) and VISKD (S4) further improve stability from the textual and visual sides, respectively. The full PAD (S5) achieves the best overall balance across domains. The same trend is observed on AKA-order2 in the supplementary material.

We further observe that VA-Prompt activation similarity strongly correlates with inter-domain feature similarity ($\rho=0.77$), suggesting that prompt routing is semantically meaningful.

\subsubsection{Analysis on Textual Distillation}
We evaluate the impact of textual distillation using weak and strong parameter configurations, alongside no-distillation and vision-only variants, as shown in Table~\ref{tab:textkd_strength}. The weak setting corresponds to $\lambda_{\text{logit}}{=}0.50$, $\tau{=}0.07$, and $\gamma{=}7.0$, while the strong setting uses $\lambda_{\text{logit}}{=}1.00$, $\tau{=}0.05$, and $\gamma{=}12.0$.

These results support our design choice that TEXKD should remain weak. Since the frozen text encoder and SupCon already provide strong implicit anchoring, a stronger text-side constraint mainly over-regularizes prompt adaptation and degrades performance. More fine-grained variants are reported in the supplementary material.

To quantify semantic drift, we measure the maximum cosine similarity between image embeddings and the frozen text prototypes at the final stage. PAD consistently improves this score by about +0.09 across representative domains (e.g., Duke 0.151$\rightarrow$0.243, Market 0.188$\rightarrow$0.277, and CUHK-SYSU 0.192$\rightarrow$0.287), supporting that TEXKD and VISKD stabilize the learned visual semantics.

\begin{table}[t]
\centering
\footnotesize
\caption{Effect of textual distillation strength on AKA-order1. Weak/Strong differ only when TEXKD is enabled; for “No KD’’ and “VISKD only’’, both columns are identical.}
\label{tab:textkd_strength}
\setlength{\tabcolsep}{8pt}
\renewcommand{\arraystretch}{1.08}
\begin{tabular}{l|cc}
\hline
\multirow{2}{*}{\textbf{Method}} &
\multicolumn{2}{c}{\textbf{TEXKD Setting}} \\
\cline{2-3}
 & \textbf{Weak (mAP / R1)} & \textbf{Strong (mAP / R1)} \\
\hline
No KD            & 68.2 / 80.1 & 68.2 / 80.1 \\
TEXKD only      & 68.4 / 80.4 & 67.6 / 79.3 \\
VISKD only       & 69.9 / 80.4 & 69.9 / 80.4 \\
TEXKD + VISKD   & \textbf{70.7 / 81.0} & 69.5 / 80.2 \\
\hline
\end{tabular}
\end{table}

\begin{figure}[t]
  \centering
  \begin{subfigure}[t]{0.49\linewidth}
    \centering
    \includegraphics[width=\linewidth]{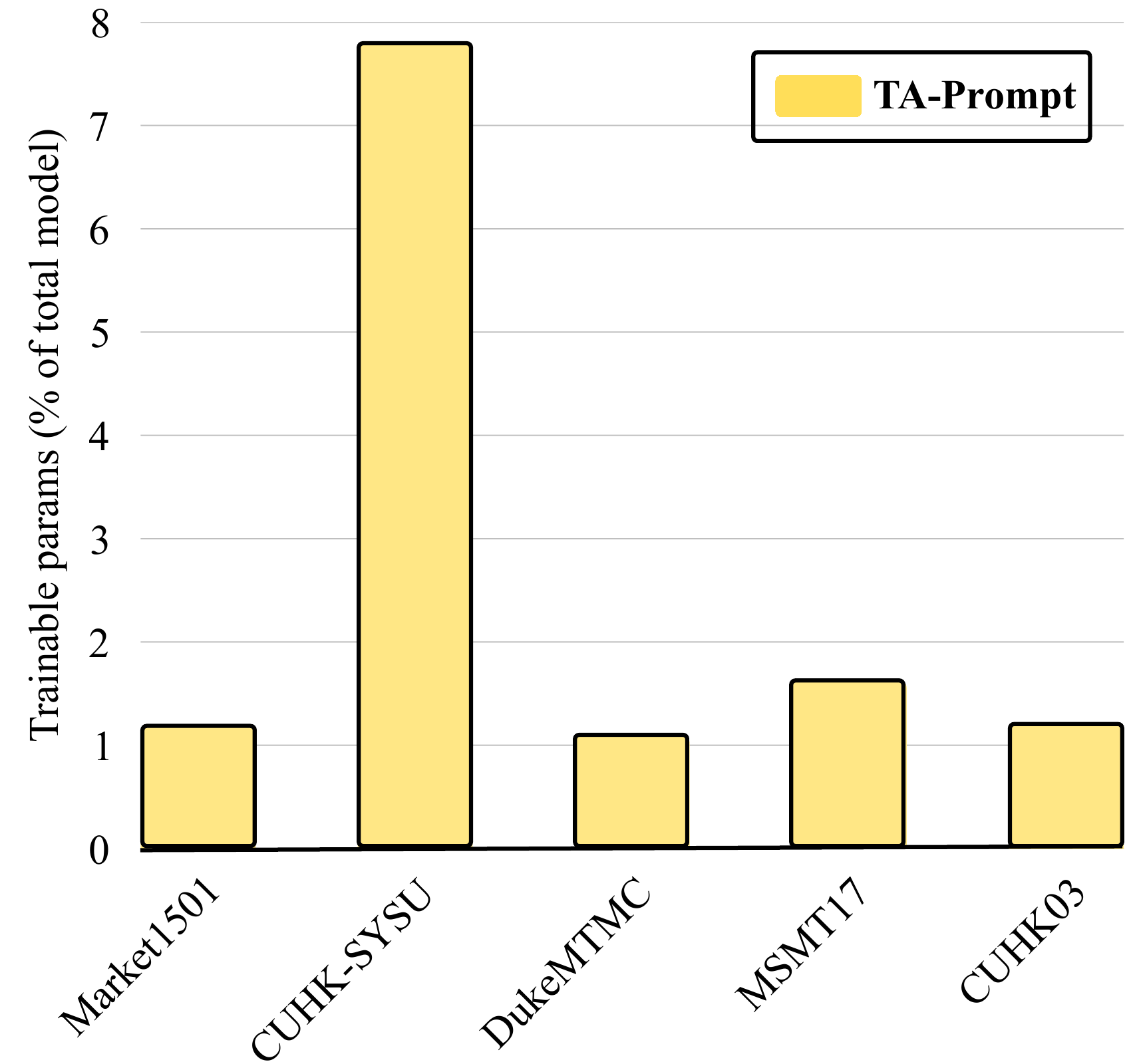}
    \caption{\textbf{TA-Prompt}.}
    \label{fig:text_side_params}
  \end{subfigure}
  \hfill
  \begin{subfigure}[t]{0.49\linewidth}
    \centering
    \includegraphics[width=\linewidth]{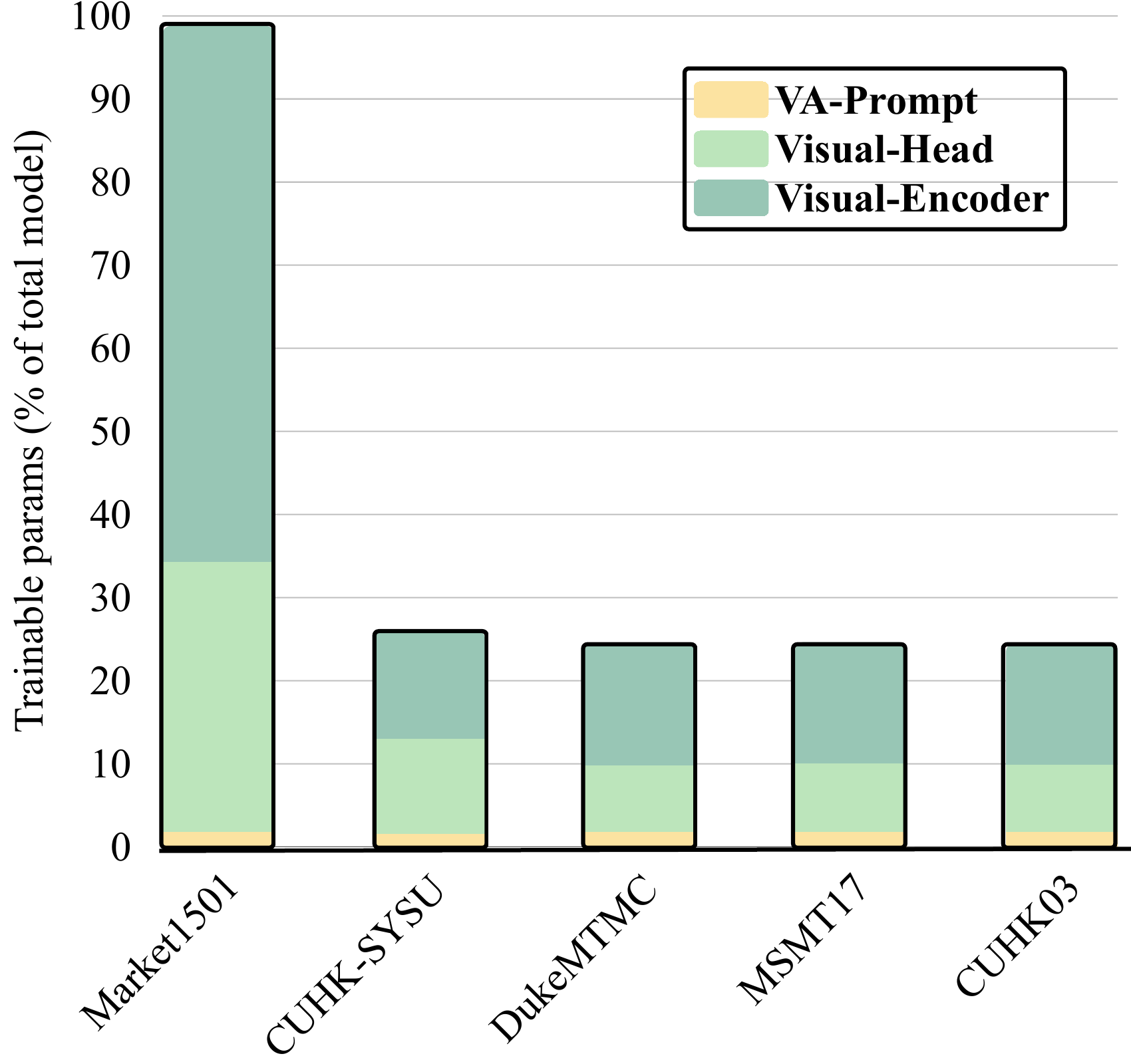}
    \caption{\textbf{VA-Prompt}, \textbf{Head}, \textbf{Backbone}.}
    \label{fig:visual_side_params}
  \end{subfigure}
  \vspace{-2mm}
  \caption{
  \textbf{Trainable parameter composition across domains.}
  The textual side \subref{fig:text_side_params} updates only the TA-Prompt, while the visual side \subref{fig:visual_side_params} trains the VA-Prompt, classifier head, and a small portion of the backbone. Ratios are normalized within trainable modules, excluding the frozen text encoder.
  }
  \label{fig:param_breakdown}
\end{figure}

\subsubsection{Parameter Freezing Strategy}
Fig.~\ref{fig:param_breakdown} summarizes the trainable ratios on both sides.
Only the TA-Prompt is updated on the textual branch, whereas the visual branch optimizes the VA-Prompt, classifier head, and selectively unfrozen backbone blocks.

Although the architecture is fixed, the effective trainable ratio varies across domains because identity distributions affect the classifier and prompt-routing parameters that receive gradients.
CUHK-SYSU, with its denser and more long-tailed identity structure, activates more visual-side parameters during Stage 1, resulting in a higher trainable ratio ($\sim$7.8\%), while other datasets stay around 1–1.6\%. 
Compared to most previous methods~\cite{zhou2025distribution,xu2024lstkc} that require full model training, our prompt-based design and selective parameter freezing drastically reduce the number of updated parameters while still enabling effective adaptation. On the 5-domain AKA benchmark, PAD introduces only marginal prompt-storage overhead, with a peak of 13.71M prompt parameters (about 26.1 MiB in FP16), and additional analysis of the number of unfrozen blocks is provided in the supplementary material.

\section{Conclusion}
\label{sec:conclusion}
We propose Prompt-Anchored vision--text Distillation (PAD) for exemplar-free lifelong person re-identification. PAD introduces a frozen textual semantic space as a persistent anchor and couples weak textual distillation with stronger EMA-guided visual distillation, so that semantic preservation and domain adaptation are handled in an asymmetric manner. In this way, PAD mitigates semantic drift without replay or full backbone fine-tuning. Extensive experiments on AKA-order1 and AKA-order2 show that PAD consistently outperforms prior methods on both seen and unseen domains. Nevertheless, some domains remain challenging, and future work may further improve PAD through adaptive distillation weighting and extensions to more complex settings such as multimodal or hybrid-clothing ReID.
\appendix
\section*{Acknowledgement}
This work was supported in part by the Natural Science Foundation of China under Grant 62402013, Grant U20B2052, Grant 61936011, in part by Grant 2023-JCJQ-LA-001-088, in part by the China Postdoctoral Science Foundation under Grant 2025T180410, Grant 2023M730056, in part by the Okawa Foundation Research Award, in part by the Ant Group Research Fund, and in part by the Kunpeng\&Ascend Center of Excellence, Peking University.
{
    \small
    \bibliographystyle{ieeenat_fullname}
    \bibliography{main}

@String(CVPR= {IEEE Conf. Comput. Vis. Pattern Recog.})

@String(ICCV= {Int. Conf. Comput. Vis.})

@String(ECCV= {Eur. Conf. Comput. Vis.})

@String(BMVC= {Brit. Mach. Vis. Conf.})

@String(ACCV  = {ACCV})

@String(AAAI = {AAAI})

@String(CVPR  = {CVPR})

@String(ICCV  = {ICCV})

@String(ECCV  = {ECCV})

@String(BMVC  =	{BMVC})

@InProceedings{li2023clip,
  title={CLIP-ReID: Exploiting Vision-Language Model for Image Re-Identification without Concrete Text Labels},
  author={Li, Siyuan and Sun, Li and Li, Qingli},
  booktitle = {AAAI},
  year={2023}
}

@inproceedings{yu2024tf,
  title={TF-CLIP: Learning Text-Free CLIP for Video-Based Person Re-identification},
  author={Yu, Chenyang and Liu, Xuehu and Wang, Yingquan and Zhang, Pingping and Lu, Huchuan},
  booktitle={AAAI},
  year={2024}
}

@inproceedings{zhai2024multi,
  title={Multi-Prompts Learning with Cross-Modal Alignment for Attribute-Based Person Re-identification},
  author={Zhai, Yajing and Zeng, Yawen and Huang, Zhiyong and Qin, Zheng and Jin, Xin and Cao, Da},
  booktitle={AAAI},
  year={2024}
}

@inproceedings{yang2024pedestrian,
  title={A pedestrian is worth one prompt: Towards language guidance person re-identification},
  author={Yang, Zexian and Wu, Dayan and Wu, Chenming and Lin, Zheng and Gu, Jingzi and Wang, Weiping},
  booktitle={Proceedings of the IEEE/CVF Conference on Computer Vision and Pattern Recognition},
  pages={17343--17353},
  year={2024}
}

@InProceedings{pu_cvpr2021,
author = {Pu, Nan and Chen, Wei and Liu, Yu and Bakker, Erwin M. and Lew, Michael S.},
title = {Lifelong Person Re-Identification via Adaptive Knowledge Accumulation},
booktitle = {CVPR},
year = {2021}
}

@inproceedings{Wu2021GeneralisingWF,
  title={Generalising without Forgetting for Lifelong Person Re-Identification},
  author={Guile Wu and Shaogang Gong},
  booktitle={AAAI},
  year={2021}
}

@inproceedings{yu2023lifelong,
  title={Lifelong Person Re-Identification via Knowledge Refreshing and Consolidation},
  author={Yu, Chunlin and Shi, Ye and Liu, Zimo and Gao, Shenghua and Wang, Jingya},
  booktitle={AAAI},
  year={2023}
}

@inproceedings{huang2022lifelong,
  title={Lifelong unsupervised domain adaptive person re-identification with coordinated anti-forgetting and adaptation},
  author={Huang, Zhipeng and Zhang, Zhizheng and Lan, Cuiling and Zeng, Wenjun and Chu, Peng and You, Quanzeng and Wang, Jiang and Liu, Zicheng and Zha, Zheng-jun},
  booktitle={CVPR},
  year={2022}
}

@inproceedings{wang2022learning,
  title={Learning to prompt for continual learning},
  author={Wang, Zifeng and Zhang, Zizhao and Lee, Chen-Yu and Zhang, Han and Sun, Ruoxi and Ren, Xiaoqi and Su, Guolong and Perot, Vincent and Dy, Jennifer and Pfister, Tomas},
  booktitle={CVPR},
  year={2022}
}

@inproceedings{wang2022dualprompt,
  title={Dualprompt: Complementary prompting for rehearsal-free continual learning},
  author={Wang, Zifeng and Zhang, Zizhao and Ebrahimi, Sayna and Sun, Ruoxi and Zhang, Han and Lee, Chen-Yu and Ren, Xiaoqi and Su, Guolong and Perot, Vincent and Dy, Jennifer and others},
  booktitle={ECCV},
  year={2022},
}

@inproceedings{xu2024lstkc,
  title={LSTKC: Long Short-Term Knowledge Consolidation for Lifelong Person Re-identification},
  author={Xu, Kunlun and Zou, Xu and Zhou, Jiahuan},
  booktitle={AAAI},
  year={2024}
}

@article{dosovitskiy2020image,
  title={An image is worth 16x16 words: Transformers for image recognition at scale},
  author={Dosovitskiy, Alexey and Beyer, Lucas and Kolesnikov, Alexander and Weissenborn, Dirk and Zhai, Xiaohua and Unterthiner, Thomas and Dehghani, Mostafa and Minderer, Matthias and Heigold, Georg and Gelly, Sylvain and others},
  journal={arXiv preprint arXiv:2010.11929},
  year={2020}
}

@inproceedings{Radford2021LearningTV,
  title={Learning Transferable Visual Models From Natural Language Supervision},
  author={Alec Radford and Jong Wook Kim and Chris Hallacy and Aditya Ramesh and Gabriel Goh and Sandhini Agarwal and Girish Sastry and Amanda Askell and Pamela Mishkin and Jack Clark and Gretchen Krueger and Ilya Sutskever},
  booktitle={ICML},
  year={2021},
}

@inproceedings{hong2025rainbowprompt,
  title={RainbowPrompt: Diversity-Enhanced Prompt-Evolving for Continual Learning},
  author={Hong, Kiseong and Kim, Gyeong-hyeon and Kim, Eunwoo},
  booktitle={Proceedings of the IEEE/CVF International Conference on Computer Vision},
  pages={1130--1140},
  year={2025}
}

@InProceedings{Zou_2025_CVPR,
    author    = {Zou, Xiaohan and Ma, Wenchao and Zhao, Shu},
    title     = {Learning Conditional Space-Time Prompt Distributions for Video Class-Incremental Learning},
    booktitle = {Proceedings of the IEEE/CVF Conference on Computer Vision and Pattern Recognition (CVPR)},
    month     = {June},
    year      = {2025},
    pages     = {4862-4873}
}

@inproceedings{smith2023coda,
  title={Coda-prompt: Continual decomposed attention-based prompting for rehearsal-free continual learning},
  author={Smith, James Seale and Karlinsky, Leonid and Gutta, Vyshnavi and Cascante-Bonilla, Paola and Kim, Donghyun and Arbelle, Assaf and Panda, Rameswar and Feris, Rogerio and Kira, Zsolt},
  booktitle={Proceedings of the IEEE/CVF conference on computer vision and pattern recognition},
  pages={11909--11919},
  year={2023}
}

@inproceedings{wang2023attriclip,
  title={Attriclip: A non-incremental learner for incremental knowledge learning},
  author={Wang, Runqi and Duan, Xiaoyue and Kang, Guoliang and Liu, Jianzhuang and Lin, Shaohui and Xu, Songcen and L{\"u}, Jinhu and Zhang, Baochang},
  booktitle={Proceedings of the IEEE/CVF Conference on Computer Vision and Pattern Recognition},
  pages={3654--3663},
  year={2023}
}

@article{wang2022s,
  title={S-prompts learning with pre-trained transformers: An occam’s razor for domain incremental learning},
  author={Wang, Yabin and Huang, Zhiwu and Hong, Xiaopeng},
  journal={Advances in Neural Information Processing Systems},
  volume={35},
  pages={5682--5695},
  year={2022}
}

@InProceedings{Zhou_2025_ICCV,
    author    = {Zhou, Da-Wei and Li, Kai-Wen and Ning, Jingyi and Ye, Han-Jia and Zhang, Lijun and Zhan, De-Chuan},
    title     = {External Knowledge Injection for CLIP-Based Class-Incremental Learning},
    booktitle = {Proceedings of the IEEE/CVF International Conference on Computer Vision (ICCV)},
    month     = {October},
    year      = {2025},
    pages     = {3314-3325}
}

@inproceedings{xu2025self,
  title={Self-Reinforcing Prototype Evolution with Dual-Knowledge Cooperation for Semi-Supervised Lifelong Person Re-Identification},
  author={Xu, Kunlun and Zhuo, Fan and Li, Jiangmeng and Zou, Xu and Jiahuan Zhou},
  booktitle={Proceedings of the IEEE/CVF International Conference on Computer Vision},
  year={2025}
}

@article{li2024exemplar,
  title={Exemplar-Free Lifelong Person Re-identification via Prompt-Guided Adaptive Knowledge Consolidation},
  author={Li, Qiwei and Xu, Kunlun and Peng, Yuxin and Zhou, Jiahuan},
  journal={International Journal of Computer Vision},
  pages={1--16},
  year={2024},
  publisher={Springer}
}

@inproceedings{xu2024distribution,
  title={Distribution-aware Knowledge Prototyping for Non-exemplar Lifelong Person Re-identification},
  author={Xu, Kunlun and Zou, Xu and Peng, Yuxin and Zhou, Jiahuan},
  booktitle={Proceedings of the IEEE/CVF Conference on Computer Vision and Pattern Recognition},
  pages={16604--16613},
  year={2024}
}

@article{zhou2025distribution,
  title={Distribution-Aware Knowledge Aligning and Prototyping for Non-Exemplar Lifelong Person Re-Identification},
  author={Zhou, Jiahuan and Xu, Kunlun and Zhuo, Fan and Zou, Xu and Peng, Yuxin},
  journal={IEEE Transactions on Pattern Analysis and Machine Intelligence},
  year={2025},
  publisher={IEEE}
}

@ARTICLE{chen2025tpami,
  author={Chen, Hao and Bremond, Francois and Sebe, Nicu and Zhang, Shiliang},
  journal={IEEE Transactions on Pattern Analysis and Machine Intelligence}, 
  title={Anti-Forgetting Adaptation for Unsupervised Person Re-Identification}, 
  year={2025},
  volume={47},
  number={2},
  pages={1056-1072},
  doi={10.1109/TPAMI.2024.3490777}}

@article{liu2025distribution,
  title={Distribution-aware Forgetting Compensation for Exemplar-Free Lifelong Person Re-identification},
  author={Liu, Shiben and Fan, Huijie and Wang, Qiang and Fan, Baojie and Tang, Yandong and Qu, Liangqiong},
  journal={arXiv preprint arXiv:2504.15041},
  year={2025}
}

@article{li2017learning,
  title={Learning without forgetting},
  author={Li, Zhizhong and Hoiem, Derek},
  journal={IEEE transactions on pattern analysis and machine intelligence},
  volume={40},
  number={12},
  pages={2935--2947},
  year={2017},
  publisher={IEEE}
}

@inproceedings{sun2022patch,
  title={Patch-based knowledge distillation for lifelong person re-identification},
  author={Sun, Zhicheng and Mu, Yadong},
  booktitle={Proceedings of the 30th ACM International Conference on Multimedia},
  pages={696--707},
  year={2022}
}

@inproceedings{cui2024learning,
  title={Learning continual compatible representation for re-indexing free lifelong person re-identification},
  author={Cui, Zhenyu and Zhou, Jiahuan and Wang, Xun and Zhu, Manyu and Peng, Yuxin},
  booktitle={Proceedings of the IEEE/CVF Conference on Computer Vision and Pattern Recognition},
  pages={16614--16623},
  year={2024}
}

@inproceedings{xu2025dask,
  title={Dask: Distribution rehearsing via adaptive style kernel learning for exemplar-free lifelong person re-identification},
  author={Xu, Kunlun and Jiang, Chenghao and Xiong, Peixi and Peng, Yuxin and Zhou, Jiahuan},
  booktitle={Proceedings of the AAAI Conference on Artificial Intelligence},
  volume={39},
  number={9},
  pages={8915--8923},
  year={2025}
}

@inproceedings{li2024fcs,
  title={Fcs: Feature calibration and separation for non-exemplar class incremental learning},
  author={Li, Qiwei and Peng, Yuxin and Zhou, Jiahuan},
  booktitle={Proceedings of the IEEE/CVF conference on computer vision and pattern recognition},
  pages={28495--28504},
  year={2024}
}

@article{Zheng2015ScalablePR,
  title={Scalable Person Re-identification: A Benchmark},
  author={Liang Zheng and Liyue Shen and Lu Tian and Shengjin Wang and Jingdong Wang and Qi Tian},
  journal={ICCV},
  year={2015},
}

@article{Xiao2017JointDA,
  title={Joint Detection and Identification Feature Learning for Person Search},
  author={Tong Xiao and Shuang Li and Bochao Wang and Liang Lin and Xiaogang Wang},
  journal={CVPR},
  year={2017}
}

@inproceedings{wei2018person,
  title={Person transfer gan to bridge domain gap for person re-identification},
  author={Wei, Longhui and Zhang, Shiliang and Gao, Wen and Tian, Qi},
  booktitle={CVPR},
  year={2018}
}

@inproceedings{ristani2016MTMC,
  title = {Performance Measures and a Data Set for Multi-Target, Multi-Camera Tracking},
  author = {Ristani, Ergys and Solera, Francesco and Zou, Roger and Cucchiara, Rita and Tomasi, Carlo},
  booktitle = {ECCV workshops},
  year = {2016}
}

@inproceedings{Gray2008ViewpointIP,
  title={Viewpoint Invariant Pedestrian Recognition with an Ensemble of Localized Features},
  author={Douglas Gray and Hai Tao},
  booktitle={ECCV},
  year={2008}
}

@INPROCEEDINGS{hirzer11,
  author = {Martin Hirzer and Csaba Beleznai and Peter M. Roth and Horst Bischof},
  title = {{Person Re-Identification by Descriptive and Discriminative Classification}},
  booktitle = {{Proc. Scandinavian Conference on Image Analysis (SCIA)}},
  year = {2011}
}

@inproceedings{Loy2009MulticameraAC,
  title={Multi-camera activity correlation analysis},
  author={Chen Change Loy and T. Xiang and S. Gong},
  booktitle={CVPR},
  year={2009}
}

@inproceedings{Li2012HumanRW,
  title={Human Reidentification with Transferred Metric Learning},
  author={W. Li and Rui Zhao and Xiaogang Wang},
  booktitle={ACCV},
  year={2012}
}

@article{Li2013LocallyAF,
  title={Locally Aligned Feature Transforms across Views},
  author={W. Li and Xiaogang Wang},
  journal={CVPR},
  year={2013}
}

@article{Li2014DeepReIDDF,
  title={DeepReID: Deep Filter Pairing Neural Network for Person Re-identification},
  author={Wei Li and Rui Zhao and Tong Xiao and Xiaogang Wang},
  journal={CVPR},
  year={2014}
}

@inproceedings{Zheng2009AssociatingGO,
  title={Associating Groups of People},
  author={Wei-Shi Zheng and Shaogang Gong and Tao Xiang},
  booktitle={BMVC},
  year={2009}
}

@article{Zhao2017SpindleNP,
  title={Spindle Net: Person Re-identification with Human Body Region Guided Feature Decomposition and Fusion},
  author={Haiyu Zhao and Maoqing Tian and Shuyang Sun and Jing Shao and Junjie Yan and Shuai Yi and Xiaogang Wang and Xiaoou Tang},
  journal={CVPR},
  year={2017}
}

@article{hermans2017defense,
  title={In defense of the triplet loss for person re-identification},
  author={Hermans, Alexander and Beyer, Lucas and Leibe, Bastian},
  journal={arXiv preprint arXiv:1703.07737},
  year={2017}
}

@article{khosla2020supervised,
  title={Supervised contrastive learning},
  author={Khosla, Prannay and Teterwak, Piotr and Wang, Chen and Sarna, Aaron and Tian, Yonglong and Isola, Phillip and Maschinot, Aaron and Liu, Ce and Krishnan, Dilip},
  journal={Advances in neural information processing systems},
  volume={33},
  pages={18661--18673},
  year={2020}
}
}

\appendix
\clearpage
\setcounter{page}{1}
\maketitlesupplementary
\renewcommand{\thefigure}{S\arabic{figure}}
\renewcommand{\thetable}{S\arabic{table}}
\setcounter{figure}{0}
\setcounter{table}{0}
\section*{Overview}
This supplementary document provides additional ablation studies and evaluation protocols to complement the analysis in the main paper. We first report results on an alternative lifelong protocol where DukeMTMC-reID is replaced by LPW-s2, verifying the robustness of PAD under different domain compositions. We then examine the influence of distillation strength on both the textual and visual branches. Next, we provide additional ablations on AKA-order2 and on the number of unfrozen transformer blocks. Finally, we study the VA-Prompt pool allocation strategy under a fixed capacity. All experimental settings remain identical to those in the main paper unless otherwise specified.

\section{Evaluation with LPW-s2}
\begin{table*}[!t]\small
	\centering
	\setlength{\tabcolsep}{4pt}
	\caption{Performance comparison with LReID methods on \textbf{Training Order-1}: Market-1501 → CUHK-SYSU → LPW → MSMT17 → CUHK03. The optimal and suboptimal values are highlighted in red and blue.}
	\label{tab:lpw_order1}
	\begin{tabular}{l|c|cc|cc|cc|cc|cc|cc|cc}
		\hline
		\multirow{2}{*}{\textbf{Method}} &
		\multirow{2}{*}{\textbf{Venue}} &
		\multicolumn{2}{c|}{Market-1501} &
		\multicolumn{2}{c|}{CUHK-SYSU} &
		\multicolumn{2}{c|}{LPW} &
		\multicolumn{2}{c|}{MSMT17} &
		\multicolumn{2}{c|}{CUHK03} &
		\multicolumn{2}{c|}{\textbf{Seen-Avg}} &
		\multicolumn{2}{c}{\textbf{Unseen-Avg}}\\
		\cline{3-16}
		& & mAP & R1 & mAP & R1 & mAP & R1 & mAP & R1 & mAP & R1 & mAP & R1 & mAP & R1\\
		\hline

		PatchKD~\cite{sun2022patch} & MM'22 
		& \textcolor{blue}{\textbf{71.6}} & \textcolor{blue}{\textbf{87.7}}
		& 77.0 & 79.6
		& 33.2 & 41.9
		& 7.0 & 18.5
		& 29.5 & 30.4
		& 43.7 & 51.6
		& 47.8 & 41.4 \\

		LSTKC~\cite{xu2024lstkc} & AAAI'24 
		& 57.0 & 78.6
		& 82.9 & 84.9
		& 47.2 & 58.4
		& \textcolor{blue}{\textbf{18.4}} & \textcolor{blue}{\textbf{41.1}}
		& 42.3 & 43.7
		& 49.6 & 61.3
		& 57.8 & 50.2 \\

		DKP~\cite{xu2024distribution} & CVPR'24 
		& 60.0 & 80.3
		& \textcolor{blue}{\textbf{84.1}} & \textcolor{blue}{\textbf{85.9}}
		& 46.0 & 57.9
		& 17.7 & 38.5
		& 41.0 & 41.4
		& 49.8 & 60.8
		& 57.5 & \textcolor{blue}{\textbf{50.7}} \\

		SPRED~\cite{xu2025self} & ICCV'25
		& 63.1 & 81.7
		& 83.2 & 84.8
		& \textcolor{blue}{\textbf{50.6}} & \textcolor{blue}{\textbf{60.7}}
		& 15.2 & 34.5
		& \textcolor{blue}{\textbf{48.6}} & \textcolor{blue}{\textbf{50.0}}
		& \textcolor{blue}{\textbf{52.1}} & \textcolor{blue}{\textbf{62.3}}
		& \textcolor{blue}{\textbf{58.7}} & \textcolor{blue}{\textbf{50.7}} \\

		\hline
		\textbf{PAD (Ours)} &
		&
		\textcolor{red}{\textbf{81.4}} & \textcolor{red}{\textbf{92.1}} &
		\textcolor{red}{\textbf{92.2}} & \textcolor{red}{\textbf{92.9}} &
		\textcolor{red}{\textbf{65.2}} & \textcolor{red}{\textbf{73.8}} &
		\textcolor{red}{\textbf{44.7}} & \textcolor{red}{\textbf{71.1}} &
		\textcolor{red}{\textbf{68.0}} & \textcolor{red}{\textbf{69.1}} &
		\textcolor{red}{\textbf{70.3}} & \textcolor{red}{\textbf{79.8}} &
		\textcolor{red}{\textbf{77.6}} & \textcolor{red}{\textbf{69.7}} \\
		\hline
	\end{tabular}
\end{table*}

\begin{table*}[!t]\small
	\centering
	\setlength{\tabcolsep}{4pt}
	\caption{Performance comparison with LReID methods on \textbf{Training Order-2}: LPW → MSMT17 → Market-1501 → CUHK-SYSU → CUHK03. The optimal and suboptimal values are highlighted in red and blue.}
	\label{tab:lpw_order2}
	\begin{tabular}{l|c|cc|cc|cc|cc|cc|cc|cc}
		\hline
		\multirow{2}{*}{\textbf{Method}} & \multirow{2}{*}{\textbf{Venue}}  &
		\multicolumn{2}{c|}{LPW} &
		\multicolumn{2}{c|}{MSMT17} &
		\multicolumn{2}{c|}{Market-1501} &
		\multicolumn{2}{c|}{CUHK-SYSU} &
		\multicolumn{2}{c|}{CUHK03} &
		\multicolumn{2}{c|}{\textbf{Seen-Avg}} &
		\multicolumn{2}{c}{\textbf{Unseen-Avg}}\\
		\cline{3-16}
		& & mAP & R1 & mAP & R1 & mAP & R1 & mAP & R1 & mAP & R1 & mAP & R1 & mAP & R1\\
		\hline

		PatchKD~\cite{sun2022patch} & MM'22
		& \textcolor{blue}{\textbf{58.0}} & \textcolor{blue}{\textbf{69.0}}
		& 6.3 & 16.7
		& 46.3 & 70.6
		& 75.7 & 78.5
		& 29.6 & 30.2
		& 43.2 & 53.0
		& 45.3 & 38.5 \\

		LSTKC~\cite{xu2024lstkc} & AAAI'24
		& 46.7 & 57.6
		& \textcolor{blue}{\textbf{14.9}} & \textcolor{blue}{\textbf{33.9}}
		& 56.5 & 78.0
		& 84.0 & 86.1
		& 42.1 & 43.7
		& 48.8 & 59.9
		& 57.4 & 49.5 \\

		DKP~\cite{xu2024distribution} & CVPR'24
		& 49.5 & 61.4
		& 14.1 & 32.6
		& \textcolor{blue}{\textbf{60.3}} & \textcolor{blue}{\textbf{80.6}}
		& \textcolor{blue}{\textbf{84.5}} & \textcolor{blue}{\textbf{86.4}}
		& \textcolor{blue}{\textbf{43.6}} & 43.7
		& \textcolor{blue}{\textbf{50.4}} & \textcolor{blue}{\textbf{60.9}}
		& \textcolor{blue}{\textbf{59.5}} & \textcolor{blue}{\textbf{52.4}} \\

		SPRED~\cite{xu2025self} & ICCV'25
		& 51.8 & 62.4
		& 10.4 & 26.2
		& 56.8 & 77.0
		& \textcolor{blue}{\textbf{84.5}} & 85.9
		& 42.9 & \textcolor{blue}{\textbf{43.8}}
		& 49.3 & 59.1
		& 57.1 & 49.1 \\

		\hline
		\textbf{PAD (Ours)}
		& &
		\textcolor{red}{\textbf{71.5}} & \textcolor{red}{\textbf{79.7}}
		& \textcolor{red}{\textbf{38.3}} & \textcolor{red}{\textbf{64.7}}
		& \textcolor{red}{\textbf{77.7}} & \textcolor{red}{\textbf{89.6}}
		& \textcolor{red}{\textbf{93.4}} & \textcolor{red}{\textbf{94.2}}
		& \textcolor{red}{\textbf{65.7}} & \textcolor{red}{\textbf{68.2}}
		& \textcolor{red}{\textbf{69.3}} & \textcolor{red}{\textbf{79.3}}
		& \textcolor{red}{\textbf{77.3}} & \textcolor{red}{\textbf{69.8}} \\
		\hline
	\end{tabular}
\end{table*}
To validate the effectiveness on more evaluation protocols, we also report the results of PAD under the LPW-s2~\cite{xu2025self} substitution protocol, where DukeMTMC-reID is replaced by LPW-s2 following recent practice in lifelong ReID. The rest of the configuration remains identical to the main paper. This evaluation is intended to verify that the behavior of PAD is stable when domain composition is slightly altered.

The performance on the LPW-based versions of AKA-order1 and AKA-order2 is summarized in \cref{tab:lpw_order1,tab:lpw_order2}, where the final stage average mAP/R1 on seen and unseen domains is shown for both orders.

Across both orders, PAD retains the same qualitative behavior as reported in the main paper, and the results confirm that the proposed framework is robust to reasonable data changes in the lifelong sequence.

\section{Ablation on Distillation Strength}
\subsection{Textual Distillation}
The main paper applies a weak but stable textual distillation on top of the frozen text feature bank, aiming to keep the TA-Prompt semantically anchored while leaving sufficient room for domain-specific adaptation. To study the role of this component, we vary the overall strength of textual distillation by jointly modifying the KL weight $\lambda_{\text{logit}}$, temperature $\tau$, logit scale $\gamma$, and the number of negatives, while keeping all other settings fixed.

We evaluate five configurations of textual distillation, denoted T1--T5, arranged from weak to strong.
T1 adopts a lighter KL weight $\lambda_{\text{logit}}$ and a mild logit scale $\gamma$.
T2 corresponds to the baseline setting used in the main paper.
T3 increases only $\lambda_{\text{logit}}$ while keeping the temperature $\tau$, logit scale $\gamma$, and the number of negatives per batch fixed, forming a clean ``pure-weight'' variant for isolating the effect of KL strength.
T4 strengthens the distillation by lowering $\tau$ and substantially enlarging $\gamma$, and additionally increases the negative batch size to enlarge the contrastive space, producing a sharper and more globally stable teacher signal.
T5 further increases $\lambda_{\text{logit}}$ on top of this sharper configuration while slightly reducing $\gamma$ relative to T4 (still higher than T2), and maintains the larger negative batch size, ensuring a monotonic progression of overall distillation strength.
In other words, T4 emphasizes sharper logits through a larger $\gamma$, whereas T5 emphasizes a stronger KL constraint via a larger $\lambda_{\text{logit}}$, allowing the two settings to reinforce textual distillation along complementary dimensions.
Results for both seen and unseen domains under AKA-order1 are summarized in \cref{tab:textkd_seen_unseen}, and all hyperparameters match the YAML blocks included in the main paper.

\begin{table}[t]
  \centering
  \caption{
    Effect of textual distillation strength (T1--T5).
    We report final stage average performance on both seen and unseen domains.
    Each configuration varies only in the KL weight $\lambda_{\text{logit}}$, temperature $\tau$, 
    and logit scale $\gamma$, whose values are listed in the table. 
    T1--T3 use a negative batch size of 256, while T4--T5 increase it to 512 to enlarge the contrastive space.
    \textbf{T2} is used as the baseline in the main paper.
  }
  \label{tab:textkd_seen_unseen}
  \vspace{2mm}
  \begin{tabular}{cccccccc}
    \toprule
    \multirow{2}{*}{\vspace{-1ex}ID}
    & \multirow{2}{*}{\vspace{-1ex}$\lambda$}
    & \multirow{2}{*}{\vspace{-1ex}$\tau$}
    & \multirow{2}{*}{\vspace{-1ex}$\gamma$}
    & \multicolumn{2}{c}{Seen}
    & \multicolumn{2}{c}{Unseen} \\
    \cmidrule(lr){5-6} \cmidrule(lr){7-8}
     & & & & mAP & R1 & mAP & R1 \\
    \midrule
    T1 & 0.25 & 0.07 & 4.0   & 70.7 & \textbf{81.3} & 78.5 & 71.2 \\
    \textbf{T2} & \textbf{0.50} & \textbf{0.07} & \textbf{7.0}
        & 70.7 & 81.0 & \textbf{78.6} & \textbf{71.4} \\
    T3 & 0.70 & 0.07 & 7.0   & \textbf{70.8} & \textbf{81.3} & 78.3 & 71.0 \\
    T4 & 0.70 & 0.05 & 16.0  & 70.3 & 81.0 & 78.2 & 71.1 \\
    T5 & 1.00 & 0.05 & 12.0  & 69.5 & 80.2 & 77.4 & 70.3 \\
    \bottomrule
  \end{tabular}
\end{table}

The results indicate that mild-to-moderate textual supervision (T1–T3) yields the most stable performance across stages, maintaining a consistent balance between seen and unseen domains.
As the distillation becomes increasingly sharp (T4) or overly dominant (T5), the representation begins to bias toward the teacher distribution, leading to gradual degradation in both seen and unseen performance.
These trends suggest that a weak textual constraint provides reliable trade-off between semantic alignment and continual adaptability.



\subsection{Visual Distillation}
PAD applies multi-level visual distillation through an EMA teacher, combining 
feature-level and logit-level alignment.  
The default configuration in the main paper uses a balanced weighting 
$(\lambda_{\text{feat}}, \lambda_{\text{logit}}) = (0.5, 0.5)$ with a temperature 
$\tau = 4.0$, while the EMA momentum is fixed to $0.997$ throughout training.

To assess how the strength of visual supervision influences continual adaptation, we vary the three distillation parameters $\lambda_{\text{feat}}$, $\lambda_{\text{logit}}$, and $\tau$ while keeping all other components identical. Five variants are tested, ranging from very weak to very strong, constructed by progressively increasing the distillation weights and adjusting the temperature accordingly. Performance on both seen and unseen domains under AKA-order1 is summarized in \cref{tab:viskd_seen_unseen}.

\begin{table}[t]
  \centering
  \caption{
    Effect of visual distillation strength (V1--V5).
    We report final stage average performance on both seen and unseen domains.
    Each configuration varies only in the feature- and logit-level weights 
    $(\lambda_{\text{feat}}, \lambda_{\text{logit}})$ and the temperature $\tau$.
    The EMA momentum is fixed to $0.997$ throughout training.
    \textbf{V3} corresponds to the configuration adopted in the main paper.
  }
  \label{tab:viskd_seen_unseen}
  \vspace{2mm}
  \begin{tabular}{cccccccc}
    \toprule
    \multirow{2}{*}{\vspace{-1ex}ID}
    & \multirow{2}{*}{\vspace{-1ex}$\lambda_{\text{feat}}$}
    & \multirow{2}{*}{\vspace{-1ex}$\lambda_{\text{logit}}$}
    & \multirow{2}{*}{\vspace{-1ex}$\tau$}
    & \multicolumn{2}{c}{Seen}
    & \multicolumn{2}{c}{Unseen} \\
    \cmidrule(lr){5-6} \cmidrule(lr){7-8}
     & & & & mAP & R1 & mAP & R1 \\
    \midrule
    V1 & 0.25 & 0.25 & 4.0   & 70.0 & 80.9 & 77.3 & 69.9 \\
    V2 & 0.35 & 0.35 & 4.0   & 70.2 & \textbf{81.0} & 77.7 & 70.5 \\
    \textbf{V3} & \textbf{0.50} & \textbf{0.50} & \textbf{4.0}
        & \textbf{70.7} & \textbf{81.0} & \textbf{78.6} & \textbf{71.4} \\
    V4 & 0.75 & 0.75 & 3.5   & 70.5 & 80.9 & 78.2 & 70.8 \\
    V5 & 1.00 & 1.00 & 3.0   & 70.0 & 80.3 & 77.1 & 69.7 \\
    \bottomrule
  \end{tabular}
\end{table}

The results show a clear upward trend from V1 to V3, indicating that moderate visual supervision effectively stabilizes the partially unfrozen backbone and yields consistent improvements across both seen and unseen domains.
Once the distillation weights continue to increase and the temperature is lowered (V4–V5), the gains begin to saturate and eventually reverse, with stronger variants showing mild degradation in both metrics.
This suggests that while a balanced level of visual guidance is beneficial, overly strong constraints still narrow the adaptation space and hinder the model’s ability to adjust to new domains.



\section{Component Analysis on AKA-order2}
To verify that the component-wise conclusions in the main paper are not specific to AKA-order1, we further conduct the same ablation study on AKA-order2. Following the setting in the main paper, we report four configurations: VA-Prompt only (S2), S2 + TEXKD (S3), S2 + VISKD (S4), and the full PAD model (S5). The results are summarized in Table~\ref{tab:component_order2}.

The trend is broadly consistent with AKA-order1. VA-Prompt already restores substantial plasticity, while TEXKD and VISKD further improve stability from the textual and visual sides. On AKA-order2, the gains of the two distillation pathways are not strictly monotonic, but the overall results still support their complementary roles under different lifelong orders.

\begin{table*}[t]
\centering
\small
\caption{Ablation study on AKA-order2. Columns indicate modules: \textbf{Freeze}---PAD freezing scheme, \textbf{VA}---Visual Adaptive Prompt, \textbf{TEXKD}---textual fixed distillation, \textbf{VISKD}---visual EMA distillation.}
\label{tab:component_order2}
\setlength{\tabcolsep}{4.2pt}
\renewcommand{\arraystretch}{1.05}
\begin{tabular}{lcccc|cc|cc|cc|cc|cc|cc}
\hline
\multirow{2}{*}{ID} &
\multirow{2}{*}{Freeze} &
\multirow{2}{*}{VA} &
\multirow{2}{*}{TEXKD} &
\multirow{2}{*}{VISKD} &
\multicolumn{2}{c|}{DukeMTMC} &
\multicolumn{2}{c|}{MSMT17} &
\multicolumn{2}{c|}{Market1501} &
\multicolumn{2}{c|}{CUHK-SYSU} &
\multicolumn{2}{c|}{CUHK03} &
\multicolumn{2}{c}{Seen Avg} \\
\cline{6-17}
 &  &  &  &  & mAP & R1 & mAP & R1 & mAP & R1 & mAP & R1 & mAP & R1 & mAP & R1 \\
\hline
S2 & \checkmark & \checkmark &  &  & 63.4 & 78.4 & 33.0 & 59.8 & 70.2 & 86.8 & 91.9 & 93.0 & 67.8 & 70.4 & 65.3 & 77.7 \\
S3 & \checkmark & \checkmark & \checkmark &  & 65.5 & 79.5 & 33.4 & 60.0 & 72.1 & 87.1 & 92.5 & 93.8 & \textbf{67.9} & \textbf{70.7} & 66.3 & 78.2 \\
S4 & \checkmark & \checkmark &  & \checkmark & 71.0 & \textbf{82.7} & \textbf{38.0} & \textbf{64.5} & 77.3 & 89.7 & 93.8 & 94.7 & 66.2 & 68.8 & \textbf{69.3} & \textbf{80.1} \\
S5 & \checkmark & \checkmark & \checkmark & \checkmark & \textbf{71.1} & 82.6 & 37.7 & 64.0 & \textbf{77.5} & \textbf{89.8} & \textbf{93.9} & \textbf{94.9} & 66.4 & 68.5 & \textbf{69.3} & 80.0 \\
\hline
\end{tabular}
\end{table*}

\section{Selective Layer Unfreezing}
The main paper updates the last 4 transformer blocks of the visual backbone. To justify this choice, we compare four variants that unfreeze the last 2, 4, 6, or 8 blocks while keeping all other settings fixed.

Table~\ref{tab:unfreeze_blocks} reports the final-stage seen-domain average, Market1501 performance as a proxy for early-domain retention, and the trainable parameter scale. Unfreezing more blocks slightly improves the overall average, but also reduces retention on the earliest domain and increases the training cost. Therefore, we adopt 4 blocks in the main paper as a practical balance between adaptation and efficiency.

\begin{table}[t]
  \centering
  \caption{
    Effect of the number of unfrozen blocks.
    We report the final-stage seen-domain average, Market1501 performance, and trainable parameters.
    \textbf{4 blocks} corresponds to the configuration used in the main paper.
  }
  \label{tab:unfreeze_blocks}
  \vspace{2mm}
  \setlength{\tabcolsep}{4.6pt}
  \begin{tabular}{cccccccc}
    \toprule
    \multirow{2}{*}{Blocks}
    & \multicolumn{2}{c}{Seen}
    & \multicolumn{2}{c}{Market1501}
    & \multicolumn{2}{c}{Trainable} \\
    \cmidrule(lr){2-3} \cmidrule(lr){4-5} \cmidrule(lr){6-7}
    & mAP & R1 & mAP & R1 & Param & Ratio \\
    \midrule
    2 & 68.7 & 79.3 & \textbf{82.0} & 91.8 & 23.6M & 16.29\% \\
    \textbf{4} & 70.7 & 81.0 & 81.2 & \textbf{92.0} & 37.8M & 26.07\% \\
    6 & 71.9 & 82.3 & 77.5 & 90.0 & 52.0M & 35.84\% \\
    8 & \textbf{72.0} & \textbf{82.6} & 74.3 & 88.3 & 66.2M & 45.61\% \\
    \bottomrule
  \end{tabular}
\end{table}

\section{Ablation on VA-Prompt Pool Allocation}
The VA-Prompt module maintains a small pool of expert prompts, and the main paper adopts the allocation $[8,8,8,8,4]$, which reflects a combined consideration of dataset size and the ordering of domains.  
To assess the role of this allocation strategy, we keep the global prompts and the Top-K routing fixed, and vary only the distribution of expert slots.

We evaluate four alternative layouts: a uniformly small configuration with reduced overall capacity; a uniform-full design that preserves the total capacity of the default but distributes it evenly; a head-heavy variant that allocates more slots to earlier domains; and a tail-heavy version that emphasizes later domains. The final stage average performance on seen and unseen domains under AKA-order1 is summarized in \cref{tab:vaprompt_seen}.

\begin{table}[t]
  \centering
  \caption{
    Ablation on VA-Prompt pool allocation (P1--P5).
    We report final stage average performance on both seen and unseen domains.
    Each variant modifies only the allocation of expert prompt slots, while the global prompt remains unchanged. \textbf{P1} corresponds to the default configuration used in the main paper.
  }
  \label{tab:vaprompt_seen}
  \vspace{2mm}
  \begin{tabular}{lccccccc}
    \toprule
    \multirow{2}{*}{\vspace{-1ex}ID}
    & \multirow{2}{*}{\vspace{-1ex}Slot}
    & \multirow{2}{*}{\vspace{-1ex}Total}
    & \multicolumn{2}{c}{Seen}
    & \multicolumn{2}{c}{Unseen} \\
    \cmidrule(lr){4-5} \cmidrule(lr){6-7}
     & & & mAP & R1 & mAP & R1 \\
    \midrule
    \textbf{P1} & \textbf{[8,8,8,8,4]}  & \textbf{36} & \textbf{70.7} & 81.0 & \textbf{78.6} & \textbf{71.4} \\
    P2 & [4,4,4,4,4]  & 20 & 68.7 & 80.0 & 76.5 & 69.0 \\
    P3 & [8,7,7,7,7]  & 36 & 70.2 & 80.7 & 77.8 & 70.5 \\
    P4 & [12,8,8,4,4] & 36 & 70.5 & \textbf{81.2} & 77.7 & 70.2 \\
    P5 & [4,4,8,8,12] & 36 & 70.5 & \textbf{81.2} & 77.3 & 70.1 \\
    \bottomrule
  \end{tabular}
\end{table}

The uniform-small configuration (P2) performs clearly worse than the default, showing that the expert pool cannot be compressed too aggressively without harming sequential performance.
Among capacity-matched variants (P3–P5), the uniform-full layout remains slightly weaker than the default, indicating that a balanced but non-uniform assignment is more effective than an equal split.
Both the head-heavy and tail-heavy allocations yield comparable but consistently lower results, suggesting that concentrating capacity at either end of the domain sequence provides no advantage.
Overall, the results indicate that PAD is not overly sensitive to reasonable capacity-matched allocations, while overly compressed or strongly imbalanced assignments lead to inferior performance.



\end{document}